\documentclass[conference]{IEEEtran}  % Comment this line out if you need a4paper

\IEEEoverridecommandlockouts                              % This command is only needed if 
\input{bib.def.short}

\author{Constantin Selzer$^{1}$ and Fabian B. Flohr$^{1}$% <-this % stops a space
\thanks{$^{1}$ Department of Electrical Engineering and Information Technology, Intelligent Vehicles Lab (IVL), Munich University of Applied Science, Lothstraße 34, 80335 Munich, German {\tt\small Constantin.Selzer@hm.edu}}%
}
\usepackage{amsmath,amssymb,amsfonts}
\usepackage{subcaption}
\usepackage[table,xcdraw]{xcolor} % Add the xcolor package
\usepackage{hyperref}
\def\BibTeX{{\rm B\kern-.05em{\sc i\kern-.025em b}\kern-.08em
    T\kern-.1667em\lower.7ex\hbox{E}\kern-.125emX}}

\usepackage{tikz}
\usetikzlibrary{positioning, fit, backgrounds, shapes, arrows.meta, calc, backgrounds}

\begin{document}

\newcommand{\TODO}[1]{{\textcolor{red}{{#1}}}} % suggestions by [] 
\newcommand{\mycomment}[1]{}

\title{PlanTRansformer: Unified Prediction and Planning with Goal-conditioned Transformer\\}

\maketitle

\begin{abstract}
Trajectory prediction and planning are fundamental yet disconnected components in autonomous driving. Prediction models forecast surrounding agent motion under unknown intentions, producing multimodal distributions, while planning assumes known ego objectives and generates deterministic trajectories. This mismatch creates a critical bottleneck: prediction lacks supervision for agent intentions, while planning requires this information. Existing prediction models, despite strong benchmarking performance, often remain disconnected from planning constraints such as collision avoidance and dynamic feasibility. We introduce Plan TRansformer (PTR), a unified Gaussian Mixture Transformer framework integrating goal-conditioned prediction, dynamic feasibility, interaction awareness, and lane-level topology reasoning. A teacher-student training strategy progressively masks surrounding agent commands during training to align with inference conditions where agent intentions are unavailable. PTR achieves 4.3\%/3.5\% improvement in marginal/joint mAP compared to the baseline Motion Transformer (MTR)~\cite{MTR2022} and 15.5\% planning error reduction at 5s horizon compared to GameFormer~\cite{GameFormer2023}. The architecture-agnostic design enables application to diverse Transformer-based prediction models. Project Website: \href{https://github.com/SelzerConst/PlanTRansformer}{https://github.com/SelzerConst/PlanTRansformer}
\end{abstract}

% \begin{IEEEkeywords}
%     Autonomous Driving, Dataset, Benchmark, Planning, Prediction
% \end{IEEEkeywords}

\section{Introduction}

Trajectory prediction and planning are critical yet distinct components in autonomous driving pipelines. Prediction models forecast surrounding agent motion to enable informed decision-making, while planning generates safe, feasible ego-trajectories given high-level navigation objectives. Despite operating on similar input representations, these tasks differ fundamentally in their formulations, optimization objectives, and available supervision.

Prediction assumes unknown agent intentions, necessitating multimodal trajectory distributions to capture behavioral uncertainty~\cite{MTR2022, SceneTransformer2022}. This formulation enables learning from large-scale observational datasets such as Waymo Open Motion Dataset~\cite{WAYMO2021}, nuScenes~\cite{nuScenes2020}, and Argoverse~\cite{Argoverse22023}, where annotations require only trajectory sequences without explicit intent labels. In contrast, planning formulates trajectory generation as single-mode optimization under the assumption of known ego-vehicle intentions. Current planning benchmarks such as NuPlan~\cite{NuPlan2021} provide ego focused trajectory data; however, the intentions of surrounding agents remain unlabeled during offline collection. This asymmetry highlights two complementary challenges: prediction has abundant but sparsely-annotated trajectory data (no intent labels required), while planning faces both data scarcity and missing intent supervision for surrounding agents across limited prediction horizons constrained by occlusion. DeepUrban~\cite{DeepUrban2024} demonstrates the value of occlusion-free perspectives for enabling longer trajectory horizons, yet explicit intent annotation remains a fundamental challenge across all datasets.

\begin{figure}
    \centering
    \includegraphics[width=\linewidth]{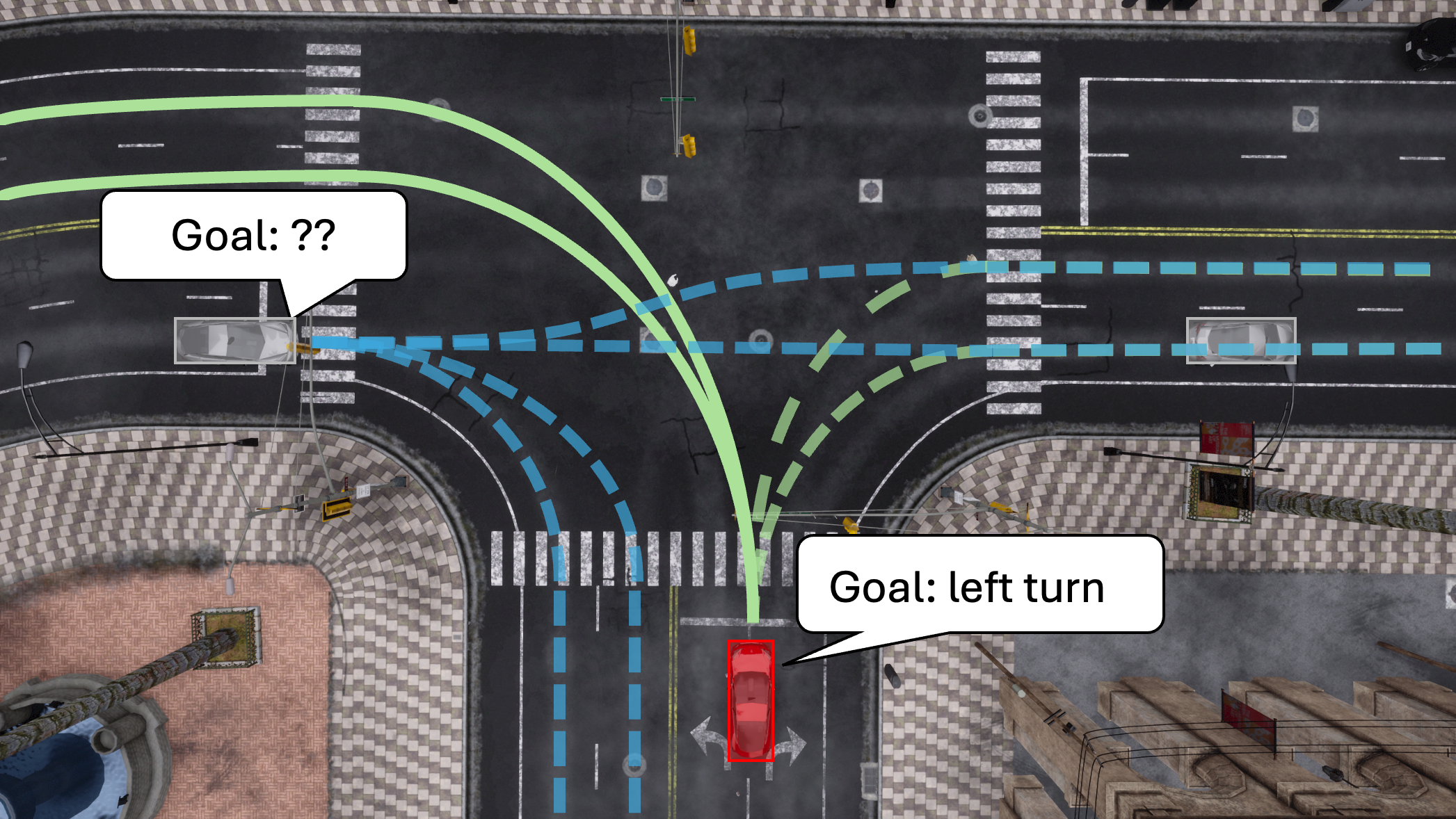}
    \caption{PTR extends prediction models with planning constrains and goal conditioning through command guidance to bridge the prediction-planning asymmetry. Left: Surrounding agents with unknown intentions produce ambiguous multimodal predictions. Right: Ego vehicle with explicit navigation context generates deterministic, goal-aligned trajectories.}
    \label{fig:motiv}
\end{figure}

This task formulation mismatch and data scarcity have resulted in a gap between prediction and planning capabilities. Most prediction models, despite achieving high accuracy on motion forecasting benchmarks, remain disconnected from planning-specific constraints such as dynamic feasibility, collision avoidance, and route consistency. The core challenge lies in bridging these two domains: incorporating planning constraints into prediction models while maintaining compatibility with existing datasets and benchmarks. Figure~\ref{fig:motiv} illustrates this prediction-planning asymmetry: surrounding agents with unknown intentions produce ambiguous multimodal predictions, whereas ego-vehicles with explicit navigation context generate deterministic, goal-aligned trajectories.

To address this gap, we propose Plan TRansformer (PTR), which extends prediction models to incorporate planning objectives and constraints through three key mechanisms: (i) differentiable dynamic feasibility and collision avoidance constraints that enforce safety during trajectory generation, (ii) goal-conditioned reasoning via high-level commands that align predictions with navigation intent, and (iii) reachable lane information being a planning element underutilized in prediction models that constrains predictions to feasible routes. Building on MTR~\cite{MTR2022}, PTR integrates these components while enforcing topological consistency with high-definition maps. Since the proposed modifications are coupled with the general Transformer architecture rather than MTR-specific design choices, our framework can be readily applied to other Transformer-based prediction models.

Our contributions are threefold. First, we systematically combine motion forecasting with planning constraints through differentiable losses, enabling joint optimization of prediction and planning objectives for safer and more feasible trajectories. Second, we propose goal-conditioning via reachable lanes and routing information, together with a teacher–student training strategy that progressively masks surrounding-agent commands to match inference conditions where agent intentions are unavailable. Third, we provide comprehensive experimental validation on the Waymo Open Motion Dataset, including detailed ablation studies that validate each component and show consistent improvements across both prediction and planning metrics.

\section{Related Work}

In the following, we review recent methods for trajectory prediction and planning.

\textbf{Data-driven Trajectory Prediction:} 
Significant advances have been made in domain-specific trajectory prediction models. A foundational model in this area is SceneTransformer~\cite{SceneTransformer2022}, which jointly predicts agent behaviors while capturing interactions between agents. By employing a masking strategy inspired by language modeling and leveraging attention mechanisms, it integrates features across agents and time, achieving strong performance in motion prediction tasks.
Further MTR~\cite{MTR2022} models trajectory prediction through joint optimization of global intention localization and local movement refinement using learnable motion query pairs for specific motion modes, improving training stability and multimodal prediction accuracy. Its agent-centered feature representation treats all agents as potential "center agents," aligning features for agents with similar future movements and circumventing transformation challenges of ego-centric methods~\cite{Su2022}.
Several extensions to MTR have been proposed. LLM-Augmented MTR~\cite{LLMMTR2024} incorporates LLMs (Large Language Models) to enhance global traffic context understanding, leveraging Transportation Context Maps and text prompts to boost prediction accuracy through a cost-efficient deployment strategy. ControlMTR~\cite{ControlMTR2024} further extends MTR by generating scene-compliant intention points and converting control commands into physics-based trajectories, which enhances road boundary adherence and improves prediction precision.
Another approach, TNT~\cite{TNT2020}, focuses on using intention points as potential goals for trajectory prediction. This method argues that leveraging such goals allows for more accurate and interpretable predictions, particularly in scenarios with multiple potential future outcomes.
Additionally, MTP2~\cite{MTP22018} emphasizes the importance of evaluating multimodal predictions through ablation studies, identifying three modes as the optimal number for achieving a balance between accuracy and computational efficiency.
Finally, VisionTrap~\cite{VisionTrap2024} enriches trajectory prediction by integrating visual input from surround-view cameras along with traditional agent tracks. This allows the model to utilize additional cues such as human gestures, road conditions, and vehicle turn signals. Moreover, textual descriptions from a VLM (Vision-Language Model) and LLM provide further guidance during training, improving the overall prediction process.

\textbf{Data-driven Trajectory Planning:}
One first approach to go from prediction to planning was PDM~\cite{PDM2023}, which extends the foundations of IDM~\cite{IDM2000} by integrating an advanced ego-forecasting component. This enhancement introduces several improvements, notably the incorporation of the GC-PGP~\cite{GC-PGP2023}, which advances the use of data-driven prediction models for trajectory planning. Another noteworthy model, Hoplan~\cite{hoplan2023}, employs a unique approach by rasterizing both current and historical trajectories of all agents onto a map, thus creating a heatmap prediction. This heatmap is then utilized by a post-solver motion planner to refine the driving trajectory for the autonomous vehicle.
In the realm of hierarchical modeling, sophisticated techniques are employed to predict and plan trajectories. GameFormer~\cite{GameFormer2023} introduces a hierarchical Transformer structure that specifically enhances interaction predictions and concentrates on the trajectory of the ego vehicle by utilizing the results from previous decoding layers. Similarly, ScePT~\cite{ScePT2022} takes a policy planning-based approach to produce trajectories that are consistent with the scene. It achieves this by segmenting the scene into interactive groups and making conditional predictions based on the dynamics within these groups.
MP3~\cite{MP32021} presents an end-to-end approach to mapless driving by processing raw sensor data and high-level commands (e.g., "turn left in 50m"). It predicts intermediate representations, including an online map and the states of dynamic agents, which are utilized by a neural motion planner to make interpretable decisions under uncertainty. Notably, MP3 employs distinct MLPs in its planning module, each tailored to execute specific high-level commands, enabling precise and context-appropriate trajectory generation.

\section{Plan TRansformer}
\begin{figure*}[h!]
    \centering
    \begin{tikzpicture}[
    scale=0.8,
  font=\small,
  image/.style={draw, rectangle, minimum width=1.5cm, minimum height=1.3cm, fill=white},
  encoder/.style={draw, rounded corners=2pt, fill=gray!20, text width=1.2cm, minimum height=1.3cm, align=center},
  circlefeat/.style={circle, draw, minimum size=0.4cm},
  rectfeat/.style={draw, rectangle, minimum width=0.4cm, minimum height=0.4cm},
  arrow/.style={-Stealth, thick}
]
%%%%%%%%%%%%%%%%%%%%%%%%%%%%%%%%%%%%%%%%%%%%%%
% Phase A: Inputs and Polyline Encoders
%%%%%%%%%%%%%%%%%%%%%%%%%%%%%%%%%%%%%%%%%%%%%%

\node[image, label={[align=center]above:Input\\Representation}, fill=red!40] (agentHist) {\includegraphics[width=0.1\linewidth]{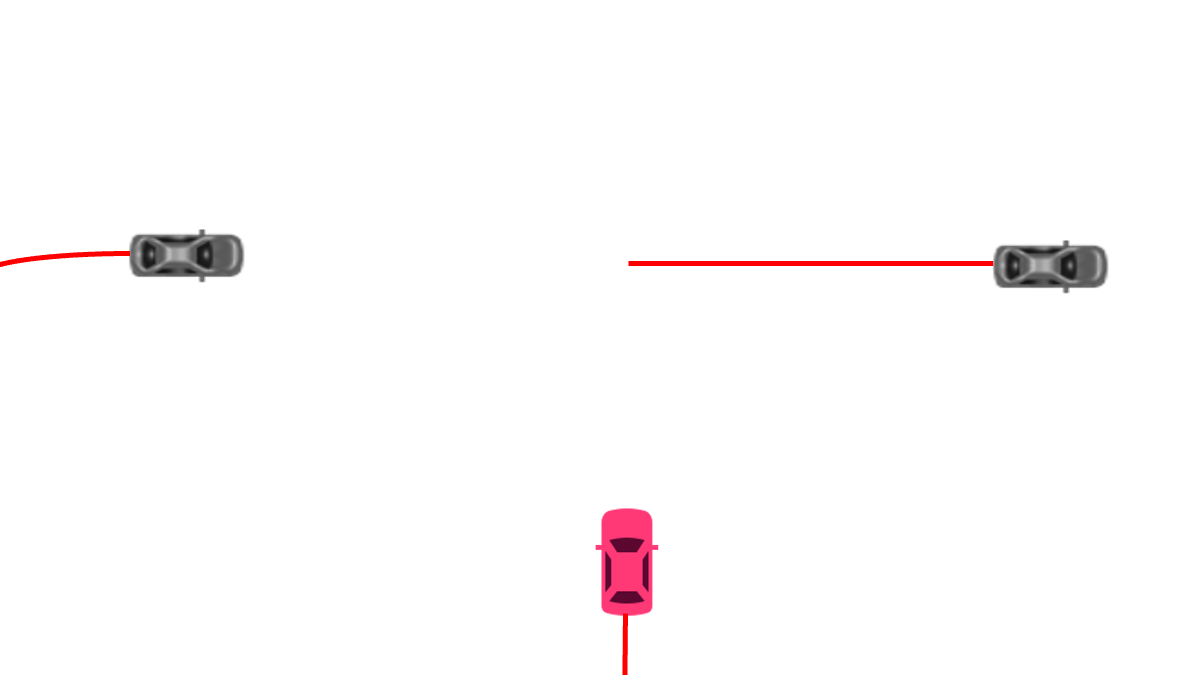}};
\node[image, label=below:Agent History $\mathbf{A}_h$, fill=red!40] (agentHist) {\includegraphics[width=0.1\linewidth]{images/Agent_history.png}};
\node[image, below=0.6cm of agentHist, label=below:Map Geometry $\mathbf{M}$, fill=green!40] (roadGeom) {\includegraphics[width=0.1\linewidth]{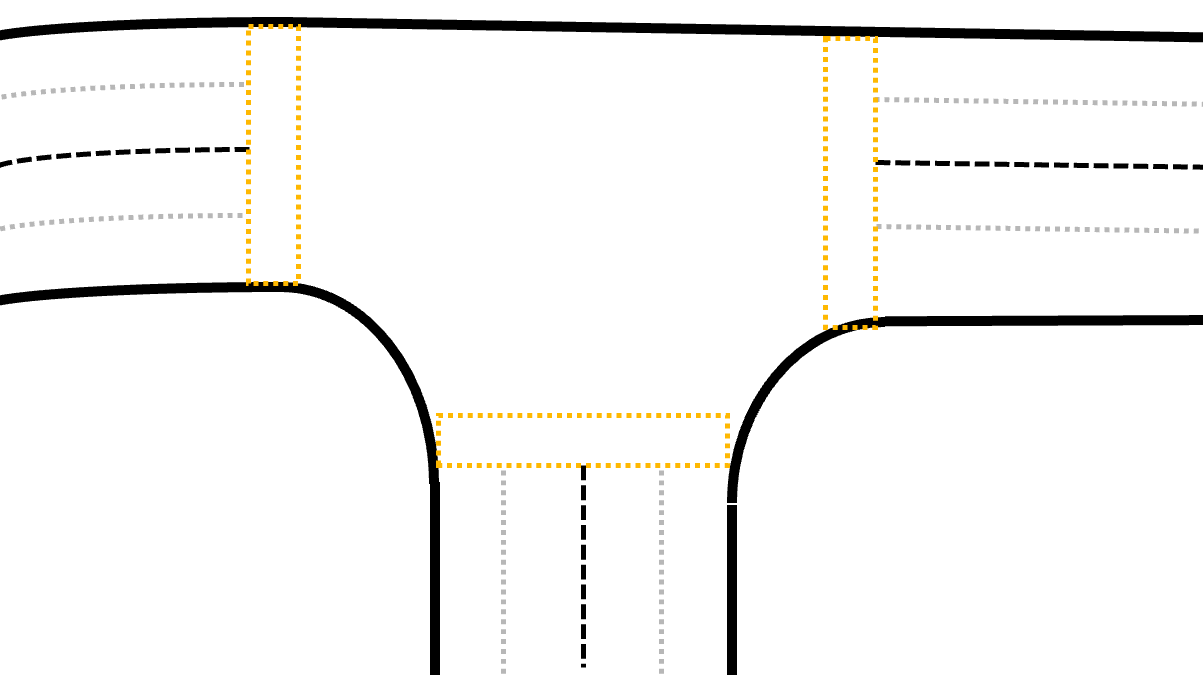}};
\node[image, below=0.6cm of roadGeom, label=below:{Route Lanes $\mathbf{L}_r$}, fill=blue!40] (reachInput) {\includegraphics[width=0.1\linewidth]{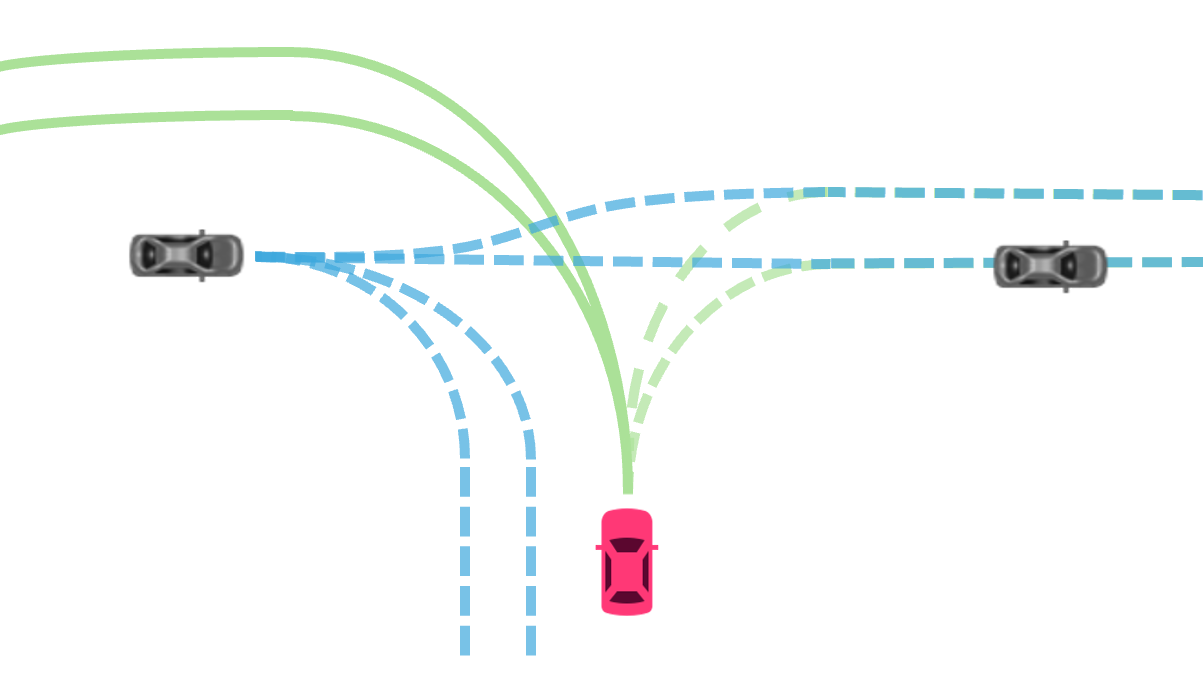}};

\node[encoder, right=0.4cm of agentHist, label={[align=center]above:Feature\\Encoding}] (enc1) {Polyline Encoder $F_A$};
\node[encoder, right=0.4cm of roadGeom] (enc2) {Polyline Encoder $F_M$};
\node[encoder, right=0.4cm of reachInput] (enc3) {Polyline Encoder $F_L$};

\draw[arrow] (agentHist) -- (enc1);
\draw[arrow] (roadGeom) -- (enc2);
\draw[arrow] (reachInput) -- (enc3);

%%%%%%%%%%%%%%%%%%%%%%%%%%%%%%%%%%%%%%%%%%%%%%
% Circle feature stacks after encoders
%%%%%%%%%%%%%%%%%%%%%%%%%%%%%%%%%%%%%%%%%%%%%%

% Align top position for all 3 stacks
\coordinate (cstackBase) at ($(enc1.east)+(0.8cm,0.7cm)$);

% === Red (Agent History) ===
\node[circlefeat, fill=red!40] (a0) at (cstackBase) {};
\node[circlefeat, fill=red!40] at ($(a0)+(0,-0.55cm)$) {};
\node at ($(a0)+(0,-1cm)$) {\dots};
\node[circlefeat, fill=red!40] at ($(a0)+(0,-1.4cm)$) {};

% === Green (Road Geometry) ===
\coordinate (a1) at ($(a0)+(0,-2.4cm)$);
\node[circlefeat, fill=green!40] at (a1) {};
\node[circlefeat, fill=green!40] at ($(a1)+(0,-0.55cm)$) {};
\node at ($(a1)+(0,-1cm)$) {\dots};
\node[circlefeat, fill=green!40] at ($(a1)+(0,-1.4cm)$) {};

% === Blue (Reachable Lanes) ===
\coordinate (a2) at ($(a1)+(0,-2.5cm)$);
\node[circlefeat, fill=blue!40] at (a2) {};
\node[circlefeat, fill=blue!40] at ($(a2)+(0,-0.55cm)$) {};
\node at ($(a2)+(0,-1cm)$) {\dots};
\node[circlefeat, fill=blue!40] at ($(a2)+(0,-1.4cm)$) {};

% Arrow from encoders to stacks
\draw[arrow] (enc1.east) -- ($(a0)+(-0.3,-0.73)$);
\draw[arrow] (enc2.east) -- ($(a1)+(-0.3,-0.7)$);
\draw[arrow] (enc3.east) -- ($(a2)+(-0.3,-0.6)$);

% === Solid box enclosing all 3 stacks ===
\node[draw, minimum width=0.5cm, minimum height=5.8cm] (circleBox) at ($(a0)+(0,-3.15)$) {};

%%%%%%%%%%%%%%%%%%%%%%%%%%%%%%%%%%%%%%%%%%%%%%
% Scene Context Encoder Box (WIDER & arrows adjusted)
%%%%%%%%%%%%%%%%%%%%%%%%%%%%%%%%%%%%%%%%%%%%%%

\path let \p1 = ($(a2)+(-0.2,-1.1cm)$) in coordinate (bBottom) at (\x1,\y1);

\node[draw, dashed, thick, fit={($(a0)+(4,-1.3)$) ($(bBottom)+(4,1.0)$)}, inner xsep=2.4cm, label=above:Scene Context Encoder] (sceneBox) {};

\node at (8.25,-2.6) {\includegraphics[width=0.25\linewidth]{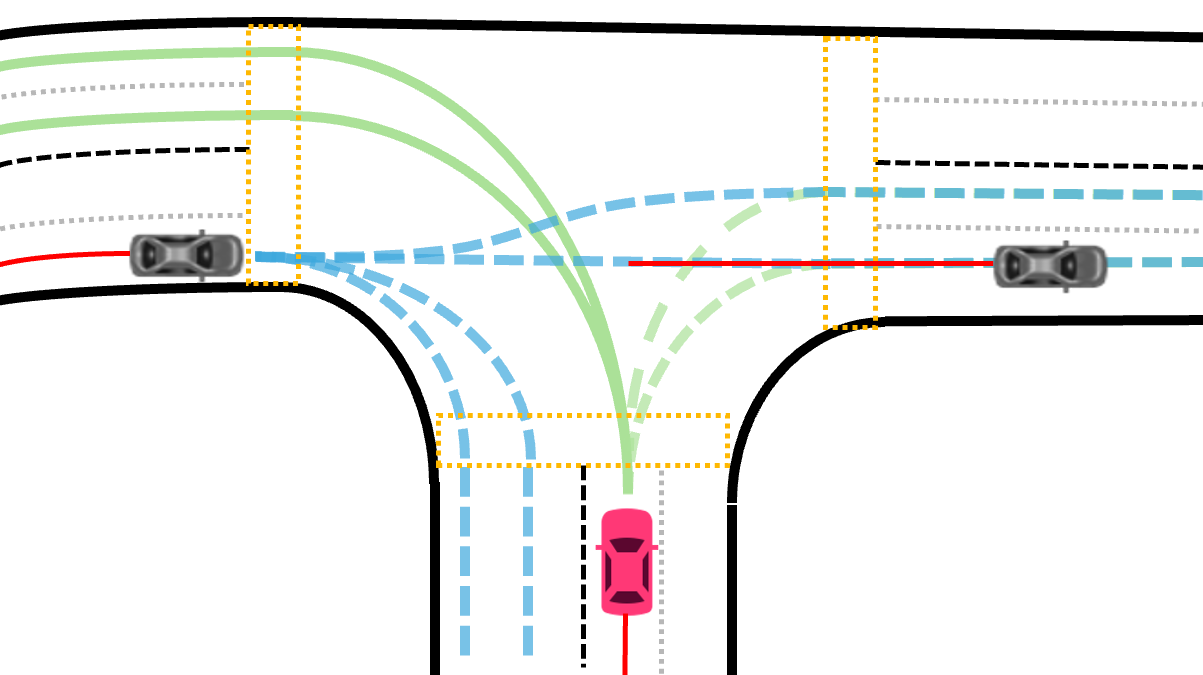}};
\node at (10,-3.6) {Local-};
\node at (10,-3.95) {Attention};
\node[] at (5.5,-5.8) {$F_{AML}$};
\node[] at (11.2,-5.8) {$F'_{AML}$};

% Output stacks for red, green, blue
\coordinate (sceneOut1) at ($(sceneBox.east)+(0.8cm,3.2cm)$);
\node[circlefeat, fill=red!40] (s0) at (sceneOut1) {};
\node[circlefeat, fill=red!40] at ($(s0)+(0,-0.55cm)$) {};
\node at ($(s0)+(0,-1cm)$) {\dots};
\node[circlefeat, fill=red!40] (s0Last) at ($(s0)+(0,-1.4cm)$) {};

\coordinate (sceneOut2) at ($(sceneBox.east)+(0.8cm,0.8cm)$);
\node[circlefeat, fill=green!40] (s1) at (sceneOut2) {};
\node[circlefeat, fill=green!40] at ($(s1)+(0,-0.55cm)$) {};
\node at ($(s1)+(0,-1cm)$) {\dots};
\node[circlefeat, fill=green!40] (s1Last) at ($(s1)+(0,-1.4cm)$) {};

\coordinate (sceneOut3) at ($(sceneBox.east)+(0.8cm,-1.6cm)$);
\node[circlefeat, fill=blue!40] (s2) at (sceneOut3) {};
\node[circlefeat, fill=blue!40] at ($(s2)+(0,-0.55cm)$) {};
\node at ($(s2)+(0,-1cm)$) {\dots};
\node[circlefeat, fill=blue!40] (s2Last) at ($(s2)+(0,-1.4cm)$) {};

% === Solid box enclosing all 3 stacks ===
\node[draw, minimum width=0.5cm, minimum height=5.8cm] (circleBox) at ($(s0)+(0,-3.15)$) {};

% Arrows from stacks to context encoder
\draw[arrow] ($(a1)+(0.3,-0.7)$) -- ($(sceneBox.west)+(0,0.05cm)$);
% Arrows from context encoder to stacks
\draw[arrow] ($(sceneBox.east)+(0,0.05cm)$) -- ($(s1)+(-0.3,-0.75)$);

%%%%%%%%%%%%%%%%%%%%%%%%%%%%%%%%%%%%%%%%%%%%%%
% Motion Decoder Box with detailed flow
%%%%%%%%%%%%%%%%%%%%%%%%%%%%%%%%%%%%%%%%%%%%%%

\node[draw, dashed, thick, fill=gray!5, minimum width=5.5cm, minimum height=4.7cm,
      right=1.5cm of sceneBox.east, anchor=west, yshift=0.6cm, label=above:Motion Decoder] (motionBox) {};

% Intention Encoder (bottom left inside box)
\node[draw, fill=gray!20, text width=2.2cm, minimum height=1cm, anchor=west]
      (dio) at ($(motionBox.west)+(0.6cm,-3.75cm)$) {Command Intention Points};

% Intermediate rectangles (horizontal stack after IQE)
\coordinate (rect1Base) at ($(dio.north)+(-0.94cm,0.9cm)$);
\node[rectfeat, fill=gray!30] (q0) at (rect1Base) {};
\node[rectfeat, fill=yellow!40] at ($(q0)+(0.6cm,0)$) {};
\node at ($(q0)+(1.3cm,0)$) {\dots};
\node[rectfeat, fill=green!40] (qLast) at ($(q0)+(2.1cm,0)$) {};

% Outer solid box
\node[draw, minimum width=2.5cm, minimum height=0.5cm] (circleBox) at ($(q0)+(1,0)$) {};

\draw[arrow] (dio.north) -- ($(q0)+(0.94cm,-0.3cm)$);

% Transformer (above Intention encoder)
\node[draw, fill=gray!20, text width=2cm, minimum height=1.5cm, anchor=west, align=center]
      (transformer) at ($(motionBox.west)+(0.6cm,-0.35)$) {Transformer Motion Decoder};

% Arrow from intermediate rectangles → transformer
\draw[arrow] ($(qLast)+(-1.2cm,0.3cm)$) -- ($(transformer.south)+(0.08,0)$);

% Final rectangles output after transformer
\coordinate (rect2Base) at ($(transformer.north)+(-1cm,0.9cm)$);
\node[rectfeat, fill=gray!30] (m0) at (rect2Base) {};
\node[rectfeat, fill=yellow!40] at ($(m0)+(0.6cm,0)$) {};
\node at ($(m0)+(1.3cm,0)$) {\dots};
\node[rectfeat, fill=green!40] (mLast) at ($(m0)+(2.1cm,0)$) {};

% Outer solid box
\node[draw, minimum width=2.5cm, minimum height=0.5cm] (circleBox) at ($(m0)+(1,0)$) {};

\draw[arrow] (transformer.north) -- ($(m0)+(1cm,-0.3cm)$);

% Label: Intention Query (from final output)
\node[font=\footnotesize] at ($(mLast)+(-1cm,1.0cm)$) {Command Intention Query};
\draw[arrow] ($(mLast)+(-1.1cm,0.4cm)$) -- ($(mLast)+(-1.1cm,0.8cm)$);

% Multimodal Prediction box (inside decoder)
\node[draw, fill=white, thick, text width=1.7cm, minimum height=1cm, anchor=west, align=center]
      (multiBox) at ($(mLast)+(1.0cm,0cm)$) {Multimodal Prediction};

\draw[arrow] ($(mLast)+(0.5cm,0)$) -- (multiBox);

% Image placeholder
\node[image, below=0.7cm of multiBox, label=below:{Predicted Scene}] (predImage) {\includegraphics[width=0.118\linewidth]{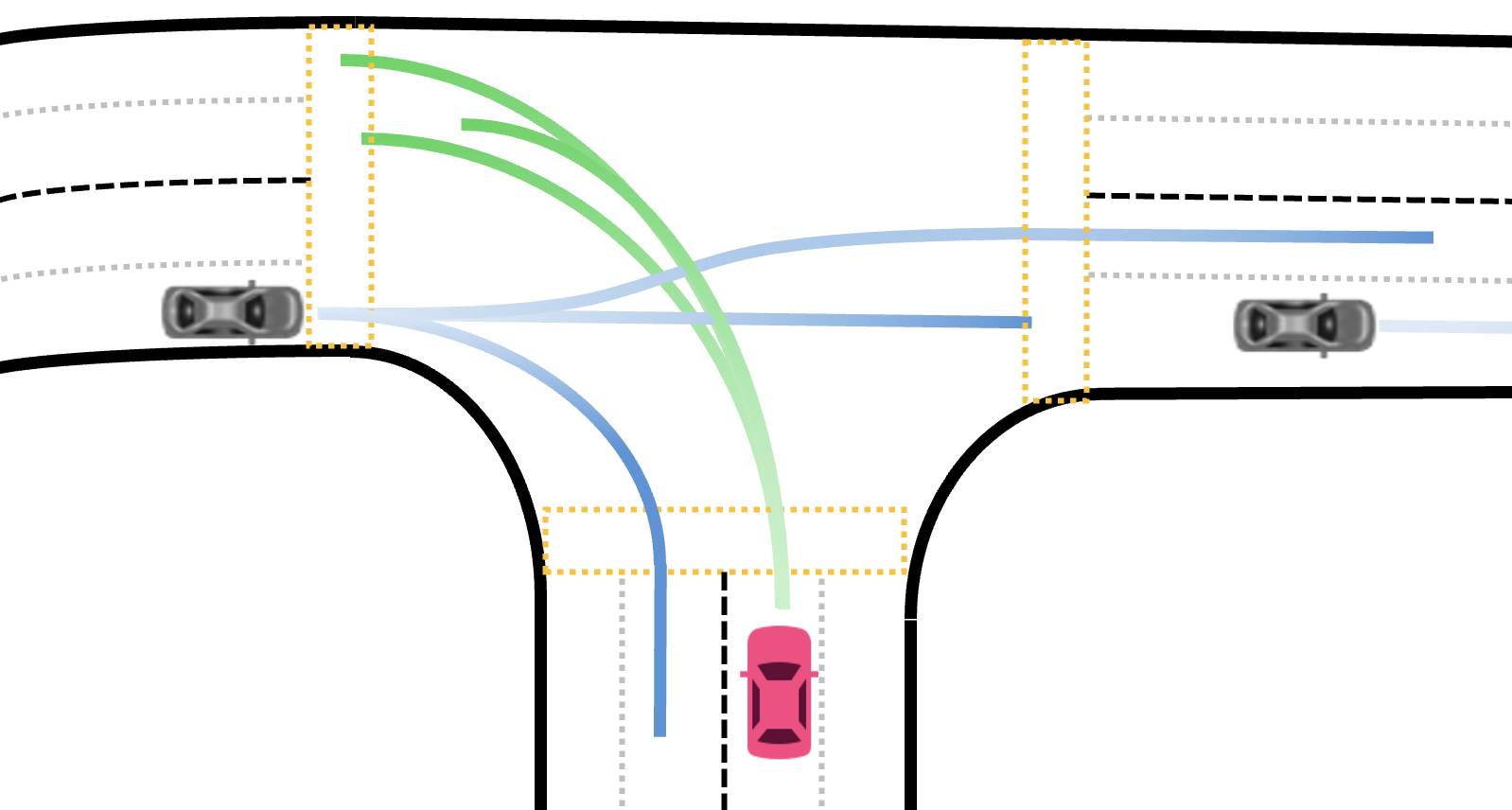}};
\draw[arrow] (multiBox.south) -- (predImage);

%%%%%%%%%%%%%%%%%%%%%%%%%%%%%%%%%%%%%%%%%%%%%%
% Scene output → Decoder input
%%%%%%%%%%%%%%%%%%%%%%%%%%%%%%%%%%%%%%%%%%%%%%
\draw[arrow] ($(s1)+(0.3cm,-0.72cm)$) -- ($(transformer.west)+(-0.2cm,-0.31cm)$);
%\draw[arrow] ($(s2Last)+(0.3,0.2)$) -- ($(dio.west)+(-0.2cm,0cm)$);
\end{tikzpicture}
    \caption{The PTR framework processes three input modalities (agent history, map polylines, reachable lanes) through polyline encoders to obtain features $F_A$, $F_M$, $F_L$. A transformer-based scene context encoder with local attention fuses these via $F_{AML}$ and decomposes into refined representations $F_A'$, $F_M'$, $F_L'$. Command-specific intention points initialize the motion decoder's query features, refined through transformer layers with scene features to produce multimodal predictions via GMM.}
    \label{fig:ptr-architecture}
\end{figure*}

\begin{figure}
\begin{tikzpicture}[
  scale=0.6,
  font=\footnotesize,
  block/.style={draw, rounded corners, minimum width=2cm, minimum height=1cm, align=center},
  token/.style={draw, rectangle, minimum width=0.3cm, minimum height=0.3cm},
  arrow/.style={-Stealth, thick},
  dashedblock/.style={draw=red!70, dashed, thick, minimum width=1.5cm, minimum height=0.8cm, font=\footnotesize, align=center},
  thinbox/.style={draw, minimum height=0.5cm, minimum width=1.8cm, font=\footnotesize, align=center},
  labelbox/.style={font=\scriptsize},
  concat/.style={draw, circle, minimum size=0.25cm, inner sep=-3pt, line width=0.8pt, font=\large, fill=white},
  circlefeat/.style={circle, draw, minimum size=0.02cm},
  image/.style={draw, rectangle, minimum width=1.5cm, minimum height=1.3cm, fill=white},
]
%%%%%%%%%%%%%%%%%%%%%%%%%%%%%%%%%%%%%%%%%%%%%%
% Features (bottom row)
%%%%%%%%%%%%%%%%%%%%%%%%%%%%%%%%%%%%%%%%%%%%%%

\node[token, fill=gray!30, xshift=-5cm] (dio0) {};
\node[token, fill=yellow!40, right=0.15cm of dio0] (dio1) {};
\node[right=0cm of dio1] (dio2) {\dots};
\node[token, fill=green!40, right=0cm of dio2] (dio3) {};
\node at ($(dio0)+(-4cm,0)$) [labelbox, text width=1.6cm] {Command Intention Points};

% === Gray dashed Motion Decoder box ===
\node[draw=gray!60!black, fill=gray!5, dashed, thick, fit={(dio0) ($(dio3)+(6.7cm,13.4cm)$)}, yshift=1.1cm, xshift=-0.8cm, inner xsep=0.8cm, inner ysep=0.4cm] (decoderBox) {};

% Outer solid box
\node[draw, minimum width=2cm, minimum height=0.5cm] (circleBox) at ($(dio0)+(0.95,0)$) {};

%%%%%%%%%%%%%%%%%%%%%%%%%%%%%%%%%%%%%%%%%%%%%%
% Initialization box
%%%%%%%%%%%%%%%%%%%%%%%%%%%%%%%%%%%%%%%%%%%%%%

\node[labelbox, above=0cm of dio1, xshift=2cm] (init) {Initialization};

%%%%%%%%%%%%%%%%%%%%%%%%%%%%%%%%%%%%%%%%%%%%%%
% Zero token
%%%%%%%%%%%%%%%%%%%%%%%%%%%%%%%%%%%%%%%%%%%%%%

\node[token, fill=white, draw=black] (zero) at ($(dio3)+(3.7cm,0)$) {};
\node at ($(zero)+(0,-0.6cm)$) [labelbox] {Command Embeding};

%%%%%%%%%%%%%%%%%%%%%%%%%%%%%%%%%%%%%%%%%%%%%%
% y^t (Dynamic) and Query modules
%%%%%%%%%%%%%%%%%%%%%%%%%%%%%%%%%%%%%%%%%%%%%%

\node[dashedblock, fill=red!5, minimum width=1.2cm, minimum height=0.6cm, above=1cm of dio1.west, xshift=-0.38cm, anchor=center] (yt) {Dynamic};
\node[labelbox, right=-1.7cm of yt] (ytt) {\(y^t\)};
\node[draw, fill=violet!30, minimum width=1.2cm, minimum height=0.6cm, above=1cm of dio2.east, xshift=0.1cm, anchor=center] (mlpstatic) {Static};

\node[labelbox, thick, text width=1.2cm, right=-3.9cm of yt] (querylabel) {Motion Query Pair};

% Arrows from Initialization
\draw[dotted, thick] ($(dio1)+(-0.88cm,0.35cm)$) -- (yt.south);
\draw[dotted, thick] ($(dio1)+(1.42cm,0.35cm)$) -- (mlpstatic.south);

%%%%%%%%%%%%%%%%%%%%%%%%%%%%%%%%%%%%%%%%%%%%%%
% Two yellow MLP + blocks above
%%%%%%%%%%%%%%%%%%%%%%%%%%%%%%%%%%%%%%%%%%%%%%

\node[draw, fill=yellow!40, minimum width=1.2cm, font=\tiny, minimum height=0.5cm, above=0.4cm of yt] (mlpyt) {Sine + MLP};
\node[draw, fill=yellow!40, minimum width=1.2cm, font=\tiny, minimum height=0.5cm, above=0.4cm of mlpstatic] (mlpsin) {Sine + MLP};

% Arrows from yt/query to yellow MLP blocks
\draw[arrow] (yt.north) -- (mlpyt.south);
\draw[arrow] (mlpstatic.north) -- (mlpsin.south);

% === Query content features (blue tokens at bottom right) ===
\coordinate (queryBase) at ($(mlpstatic)+(2.4cm,0cm)$);
\foreach \i in {0,1,2,3,4} {
  \node[token, fill=blue!30] (qt\i) at ($(queryBase)+(\i*0.6cm,0)$) {};
}
% Outer solid box
\node[draw, minimum width=2.0cm, minimum height=0.5cm, xshift=-0.2cm] (queryBox) at ($(qt0)+(1.5,0)$) {};

\node at ($(queryBase)+(4cm,0)$) [labelbox, text width=1cm] {\(C^{t} \)};

\node at ($(queryBase)+(4cm,-1.42cm)$) [labelbox, text width=1cm] {Query Content Feature};

\draw[arrow] (zero) -- ($(qt2)+(0.2,-0.4)$);

% === Green dashed Self-Attention box ===
\node[draw=green!60!black, fill=green!5, dashed, thick, fit={(qt0) ($(qt4)+(-0.3cm,4cm)$)}, yshift=1.2cm, xshift=0.4cm, inner xsep=0.8cm, inner ysep=0.4cm] (selfBox) {};
\node[labelbox, text=green , above=0cm of selfBox, xshift=0.7cm] (selflabel) {\footnotesize Self-Attention};

% Multi-Head Attention inside green box
\node[draw, fill=orange!40, text width=2cm, minimum height=0.8cm, align=center, above=2.1cm of queryBase, xshift=0.8cm] (mhattn) {Multi Head Attention};
\node[draw, text width=2cm, fill=white, minimum height=0.4cm, align=center, above=0.32cm of mhattn] (addnorm) {Add \& Norm};

\node[concat] (concat1) at ($(mlpsin)+(2.7cm,0cm)$) {+};

% Arrows from tokens to MH-Attention
\draw[arrow] ($(qt0)+(0.3cm,0.4cm)$) -- (concat1.south);
\draw[arrow] (mlpsin.east) -- ($(concat1.west)+(0,0)$);
\draw[arrow] (concat1.north) -- ($(mhattn.south)+(-1.03cm,0)$);
\draw[arrow] (concat1.east) -| (mhattn.south);
\draw[arrow] ($(qt4)+(-0.2cm,0.4cm)$) -- ($(mhattn.south)+(0.9cm,0)$);
\draw[thick] ($(qt4)+(-0.2cm,1.6cm)$) -- ($(qt4)+(1.4cm,1.6cm)$);
\draw[arrow] ($(qt4)+(1.4cm,1.6cm)$) |- (addnorm.east);

% Arrow from MH-Attn → Add & Norm
\draw[arrow] (mhattn.north) -- (addnorm.south);

% === Orange dashed Cross-Attention box ===
\node[draw=orange!60!black, fill=orange!5, dashed, thick, fit={(dio0) ($(dio3)+(0.9cm,6cm)$)}, yshift=3cm, xshift=-0.7cm, inner xsep=0.8cm, inner ysep=0.4cm] (crossBox) {};

\node[labelbox, text=orange, above=0.0cm of crossBox.north, xshift=1.3cm] (selflabel) {\footnotesize Cross-Attention};

% Output anchor for future connection
\coordinate (selfOut) at (addnorm.west);

% === Multi-Head Attention block (center left) ===
\node[draw, fill=orange!40, text width=2cm, minimum height=0.8cm,
      above=0.9cm of mlpyt, align=center, xshift=0.5cm] (mainAttn) {Multi Head Attention};

% Output stacks for red, green, blue
\coordinate (sceneOut1) at ($(mainAttn.west)+(-2.2cm,5.5cm)$);
\node[circlefeat, fill=red!40] (s0) at (sceneOut1) {};
\node[circlefeat, fill=red!40] at ($(s0)+(0,-0.6cm)$) {};
\node at ($(s0)+(0,-1cm)$)[scale=0.6, thick] {\vdots};
\node[circlefeat, fill=red!40] (s0Last) at ($(s0)+(0,-1.5cm)$) {};

\coordinate (sceneOut2) at ($(mainAttn.west)+(-2.2cm,3.4cm)$);
\node[circlefeat, fill=green!40] (s1) at (sceneOut2) {};
\node[circlefeat, fill=green!40] at ($(s1)+(0,-0.6cm)$) {};
\node at ($(s1)+(0,-1cm)$)[scale=0.6, thick] {\vdots};
\node[circlefeat, fill=green!40] (s1Last) at ($(s1)+(0,-1.5cm)$) {};

\coordinate (sceneOut3) at ($(mainAttn.west)+(-2.2cm,1.3cm)$);
\node[circlefeat, fill=blue!40] (s2) at (sceneOut3) {};
\node[circlefeat, fill=blue!40] at ($(s2)+(0,-0.6cm)$) {};
\node at ($(s2)+(0,-1cm)$)[scale=0.6, thick] {\vdots};
\node[circlefeat, fill=blue!40] (s2Last) at ($(s2)+(0,-1.5cm)$) {};

% === Solid box enclosing all 3 stacks ===
\node[draw, minimum width=0.4cm, minimum height=3.9cm] (circleBox) at ($(s0)+(0,-2.85)$) {};

\node[draw, fill=white, text width=2cm, minimum height=0.4cm, align=center, above=0.3cm of mainAttn] (mainNorm1) {Add \& Norm};

\node[labelbox, thick, right=-1.3cm of sceneOut2, text width=1cm] (contextlabel) {Context Feature};

\node[concat] (concat2) at ($(mlpyt)+(0.83cm,1.1cm)$) {+};
\node[concat] (concat3) at ($(mlpyt)+(-2cm,1.1cm)$) {+};

% Circle with sine, right of refPoint
\begin{scope}[yshift=4.35cm, xshift=-11.6cm]  % relative movement
  \draw[thick, fill=white] circle(0.3cm);
  \draw[domain=-1:1, scale=0.3, smooth, variable=\x, thick]
    plot ({\x}, {0.3*sin(180*\x)});
\end{scope}

\node[labelbox, thick, right=-2.2cm of concat3, text width=1cm] (positionlabel) {Positional Encoding};

% Arrow from yellow MLP to Concatenation 2
\draw[arrow] ($(mlpyt.north)+(0.83cm,0)$) -- (concat2.south);
% Arrow from Concatenation 2 to Multi Head Attention
\draw[arrow] (concat2.north) -- (mainAttn.south);

% Arrow from Position Encoding to Concatenation 3
\draw[arrow] ($(concat3.west)+(-0.6cm,0)$) -- (concat3.west);

% Arrow from Concatenation 3 to Multi Head Attention
\draw[arrow] (concat3.west) -| ($(mainAttn.south)+(-1cm,0)$);

% Arrow from Scene Context Encoder to Multi Head Attention
\draw[arrow] ($(s2.east)+(0.05,-1.3)$) -- (mainAttn.west);

% Arrow from Scene Context Encoder to Multi Head Attention
\draw[arrow] ($(s2.east)+(0.05,-1.3)$) -| (concat3.north);

% Arrow from Self-Attention output (Phase 2) to Add Norm 1
\draw[arrow] (selfOut) -- ($(mainNorm1.east)+(0,0cm)$);

% Arrow from Self-Attention output (Phase 2) to Concatenation 2
\draw[arrow] ($(selfOut)+(-1cm,0)$) |- ($(concat2.east)+(0,0cm)$);

% === Feedforward block (above) ===
\node[draw, fill=cyan!30, text width=2cm, minimum height=0.8cm, align=center,
      above=0.6cm of mainNorm1] (feedforward) {Feed Forward};

\node[draw, fill=white, minimum width=3cm, minimum height=0.4cm,
      above=0.3cm of feedforward] (mainNorm2) {Add \& Norm};

% Arrows within the central stack
\draw[arrow] (mainAttn.north) -- (mainNorm1.south);
\draw[arrow] (mainNorm1.north) -- (feedforward.south);
\draw[thick] ($(mainNorm1.north)+(0,0.2cm)$) -- ($(mainNorm1.north)+(-3cm,0.2cm)$);
\draw[arrow] ($(mainNorm1.north)+(-3cm,0.2cm)$) |- (mainNorm2.west);
\draw[arrow] (feedforward.north) -- (mainNorm2.south);

% === Multimodal Prediction Block ===
\node[draw, fill=gray!20, minimum width=2.8cm, minimum height=0.8cm,
      above=0.6cm of mainNorm2] (multiPred) {Multimodal Prediction};

% === GMM Prediction ===
\node[image, minimum height=1cm, minimum width=1.5cm, right=0.7cm of multiPred, label=above:GMM Prediction, fill=gray!50] (gmmPred) {\includegraphics[width=0.18\linewidth]{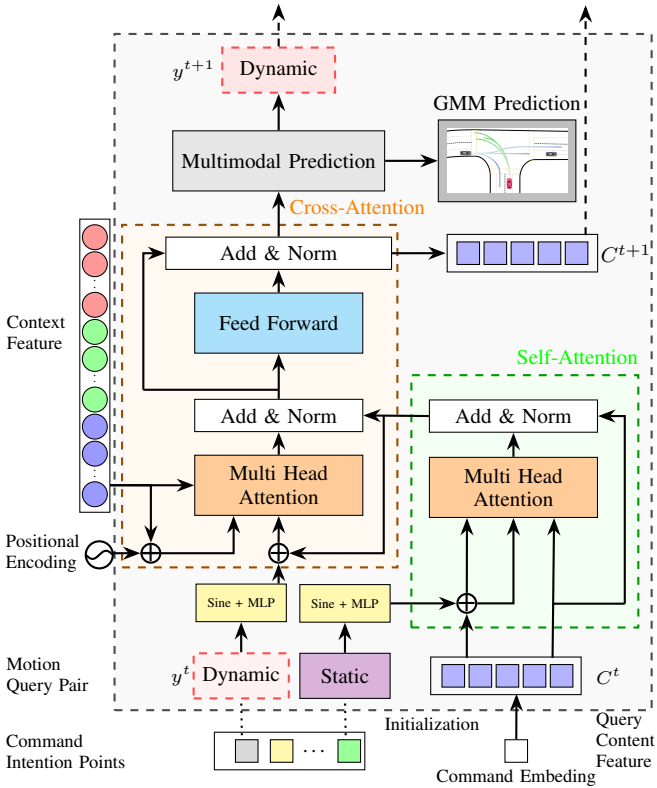}};

% Arrow from Add & Norm → Multimodal Prediction
\draw[arrow] (mainNorm2.north) -- (multiPred.south);

% Arrow from Add & Norm → Multimodal Prediction
\draw[arrow] (multiPred.east) -- (gmmPred.west);

% === Query Content (tokens) ===
\coordinate (queryBase) at ($(mainNorm2)+(4.2cm,0)$);
\foreach \i in {0,1,2,3,4} {
  \node[token, fill=blue!30] (qe\i) at ($(queryBase)+(\i*0.6cm,0)$) {};
}

% Arrow from Query Content to Outside
\draw[arrow, dashed] ($(qe4)+(0.2cm,0.4cm)$) -- ($(qe4)+(0.2cm,5.5cm)$);

% Outer solid box
\node[draw, minimum width=2cm, minimum height=0.5cm] (queryBox1) at ($(qe0)+(1.2,0)$) {};

\node at ($(qe3)+(1.7,0)$) {\footnotesize \(C^{t+1}\)};

% Arrow from Multimodal → GMM tokens
\draw[arrow] (mainNorm2.east) -- ($(qe0)+(-0.5cm,0)$);

% === y^{t+1} Dynamic (top left, dashed block) ===
\node[dashedblock, fill=red!10, minimum height=0.6cm, above=0.8cm of multiPred.north, anchor=center] (ytnext) {Dynamic};
\node[labelbox, right=-2.3cm of ytnext] (yttnext) {\(y^{t+1}\)};

\draw[arrow, dashed] (ytnext.north) -- ($(ytnext.north)+(0,0.95cm)$);

\draw[arrow] (multiPred.north) -- (ytnext.south);

% Dashed arrow back to y^{t+1}
%\draw[dashed, thick, ->] (multiPred.north west) to[out=135,in=-45] (ytnext.south);

\end{tikzpicture}
    \caption{The decoder of the PTR framework integrates command-guided query initialization with reachable lane features. Static and dynamic motion queries are initialized from high-level commands and refined iteratively through self-attention and cross-attention mechanisms with scene context and lane features, producing multimodal trajectory predictions via GMM.}
    \label{fig:ptr-decoder}
\end{figure}

PTR extends MTR to the planning domain by integrating goal-conditioned trajectory generation with real-world driving constraints. The framework employs a transformer-based encoder-decoder structure to model complex scenarios and predict diverse, rule-compliant trajectories. Figure~\ref{fig:ptr-architecture} illustrates the overall architecture, while Figure~\ref{fig:ptr-decoder} details the decoder processing of command embeddings and route features.

PTR comprises three key components: Section~\ref{sec:scene_en} a scene encoding module capturing agent history, map features, and interactions; Section~\ref{sec:future_de} a future decoding module utilizing route-based intention points and high-level commands for trajectory generation; and Section~\ref{sec:learning_proc} a learning process optimizing for trajectory accuracy and planning-specific requirements.

The following sections detail each component.

%%%%%%%%%%%%%%%%%%%%%%%%%%%%%%%%%%%%%%%%%%%%%%%%%%%%%%%%%%%%%%%

\subsection{Scene Encoding}
\label{sec:scene_en}

\subsubsection{Input Representation}

The PTR framework processes three primary input modalities (Figure~\ref{fig:ptr-architecture}, Input Representation): agent historical trajectories, vectorized map polylines, and reachable lanes. Following vectorized representation principles, all inputs are organized as polylines and normalized to agent-centric coordinates. Agent histories are represented as $\mathbf{A}_h \in \mathbb{R}^{N_{c} \times N_{a} \times T_{h} \times d_{s}}$, where $N_c$ denotes center agents, $N_a$ is total agents, $T_h$ is the historical window, and $d_s$ is the state dimensionality encompassing position, velocity, angle, and heading.
Map polylines are encoded as $\mathbf{M} \in \mathbb{R}^{N_{c} \times N_{m} \times N_{p} \times d_{p}}$, with $N_m$ map elements per center agent, $N_p$ waypoints per element, and $d_p$ waypoint dimensionality including position, orientation, and polyline type.
Reachable lanes are represented as $\mathbf{L}_r \in \mathbb{R}^{N_{c} \times N_{l} \times N_{p}^{l} \times d_{l}}$, where $N_l$ is the number of feasible destination lanes and $d_l$ captures lane attributes including position and orientation. This explicit inclusion of reachable lanes extends the MTR framework by providing navigation constraints. When navigation commands are provided, reachable lanes are filtered geometrically to further constrain the prediction space. All inputs are zero-padded for consistency.

\subsubsection{Feature Encoding}

Agent historical states are processed through temporal MLPs with max-pooling aggregation, yielding $F_A \in \mathbb{R}^{N_{c} \times N_{a} \times D}$ that captures long-term dependencies (Figure~\ref{fig:ptr-architecture}, Feature Encoding). Map polylines and reachable lanes follow a PointNet-like architecture: waypoint features are independently processed by MLPs and aggregated via max-pooling to obtain $F_M \in \mathbb{R}^{N_{c} \times N_{m} \times D}$ and $F_L \in \mathbb{R}^{N_{c} \times N_{l} \times D}$, where $D$ denotes the projected feature dimensionality. This polyline encoding strategy efficiently summarizes each geometric element as a single token feature.

\subsubsection{Scene Context Encoder}

Agent, map, and lane features are concatenated to form the fused representation $F_{AML} \in \mathbb{R}^{N_{c} \times (N_{a} + N_{m} + N_{l}) \times D}$ (Figure~\ref{fig:ptr-architecture}, Scene Context Encoder). A transformer-based encoder with local self-attention processes this concatenated feature, leveraging positional encodings $P_{AML} \in \mathbb{R}^{N_{c} \times (N_{a} + N_{m} + N_{l}) \times 2}$ from polyline centers and agent positions. The local attention mechanism maintains scene locality by restricting attention to k-nearest neighboring polylines for each query, important for modeling road map relationships while remaining memory-efficient. This locality-preserving design maintains spatial coherence essential for navigation behaviors.

Iterative attention refinement across encoder layers produces constraint-aware scene features $F_{AML}'$, decomposed into refined agent features $F_A'$, map features $F_M'$, and lane features $F_L'$ for decoder processing.

\subsection{Future Decoding}
\label{sec:future_de}

The trajectory decoder employs a transformer-based architecture with learnable motion query pairs to generate multimodal trajectories from scene and agent-specific features. We adopt MTR's decoder with two key modifications: high-level command guidance and reachable lane constraints.

\subsubsection{High-Level Command Guidance and Query Initialization}

High-level commands (HLCs) categorize agent intentions into six semantic types: left turn, straight, right turn, stationary, unknown, and  vulnerable road user (VRU). Commands are determined via rule-based heuristics from agent geometry and dynamics (displacement, heading, velocity), serving as auxiliary supervision during training and dynamically assigned at inference from planned waypoints. For ego agents, commands guide maneuver planning; non-ego agents default to unknown.
Each command type initializes queries via learned embeddings $\mathbf{e}_{c} \in \mathbb{R}^{D}$, conditioning the query content features $\mathbf{C}^{0} \in \mathbb{R}^{K \times D}$ (Figure~\ref{fig:ptr-decoder}, Initialization). This command-conditioned initialization incorporates semantic priors into query initialization, improving convergence and biasing the decoder toward intention-aligned trajectories.
For command-specific clustering, vehicle commands cluster ground-truth (GT) endpoints conditioned on their distributions, while unknown commands and VRUs use global MTR clustering to maintain coverage for unpredictable motions. VRUs require independent processing due to unconstrained motion characteristics: pedestrians move omnidirectionally, while cyclists exhibit mixed road compliance.

\subsubsection{Reachable Lane Features and Context Feature}

Reachable lanes encoded as $F_L' \in \mathbb{R}^{N_{c} \times N_{l} \times D}$ during scene encoding provide explicit navigation constraints. These refined lane features are projected into the decoder's embedding space and integrated via cross-attention alongside agent features $F_A'$ and map features $F_M'$ (see Figure~\ref{fig:ptr-decoder}), enabling the decoder to respect route feasibility and lane topology. The decoder simultaneously aggregates agent features, lane features, and map features through dedicated attention modules, progressively refining predictions based on scene-compliant navigation possibilities. Explicit modeling and refinement of reachable lane features enable route-compliant trajectory generation.

\subsubsection{Iterative Refinement and Multimodal Output}

Motion query pairs with static intention and dynamic searching queries propagate through self-attention (Figure~\ref{fig:ptr-decoder}, green dashed box) and aggregate features via cross-attention (Figure~\ref{fig:ptr-decoder}, orange dashed box) with scene context. The decoder combines: (i) scene context with agents and maps, (ii) command-guided query features, and (iii) reachable lanes. Cross-attention modules aggregate these with agent and map representations, processed through feed-forward networks with residual connections (Add \& Norm layers). Across iterations, the decoder produces multimodal trajectory distributions via Gaussian Mixture Model heads (Figure~\ref{fig:ptr-decoder}, top right).

The final trajectory prediction for agent $i$ is
\[
\hat{\mathbf{Y}}_i \in \mathbb{R}^{T_f \times 2},
\]
where $T_f$ denotes the prediction horizon, ensuring predictions remain statistically consistent with observed motion while adhering to semantic intent and navigational feasibility.

% Dynamic traffic light states may be incorporated as follows:
% During the cross-attention process applied to the decoder's context, the dynamic states of traffic lights at each lane can be incorporated into the attention mechanism. Lanes with a green signal receive increased attention weights, as they are more likely to be traversed, whereas lanes with yellow or red signals are assigned lower attention weights to reflect their reduced accessibility. For lanes where traffic light information is unavailable, the attention distribution remains unchanged. This integration ensures that the model accounts for real-time traffic regulations, further constraining trajectory predictions to comply with traffic rules and improving overall prediction fidelity in signalized environments.

\subsection{Learning Process}
\label{sec:learning_proc}
Our framework is trained end-to-end with a multi-objective loss balancing trajectory accuracy, multimodality, and safety. Following MTR, we adopt hard-assignment strategy that selects the motion query pair closest to the GT endpoint, which serves as the positive Gaussian component for optimization.

\subsubsection{Base Trajectory Losses}
We incorporate three foundation losses from prior work. The dense prediction loss \(L_{\text{dense}}\) is an \(\ell_1\) regression loss optimizing auxiliary future trajectory predictions for capturing multi-agent interactions. The Gaussian regression loss \(L_{\text{GMM}}\) applies negative log-likelihood over predicted Gaussian components, maximizing the likelihood of ground-truth positions and the probability of selected positive modes. The classification loss \(L_{\text{cls}}\) enforces cross-entropy on predicted trajectory mode probabilities. These base losses are applied uniformly across all decoder layers.

\subsubsection{Collision Loss}

The collision loss encourages socially-aware and safe trajectories by penalizing spatial proximity between agent pairs. It uses axis-wise smooth penalty functions with softplus gating, aggregated across all prediction modes and time steps to account for safety margins:

\begin{equation}
\mathcal{L}_{\text{col}} = \frac{1}{BM} \sum_{b,m,t} \rho_x \cdot \rho_y \cdot w_t \cdot \mathbb{1}(\text{valid}).
\end{equation}
where $B$ is the batch size, $M$ is the number of modes, $\rho_x$ and $\rho_y$ are the axis-wise distance penalties, $w_t$ is a time-decaying weight prioritizing near-term safety, and $\mathbb{1}(\text{valid})$ indicates valid agent pairs.

\subsubsection{Dynamics Loss}
The dynamics loss enforces kinematic feasibility by penalizing the L2 deviation between direct spatial predictions and positions simulated via integrated control outputs (yaw and speed):

\begin{equation}
\mathcal{L}_{\text{dynamic}} = \frac{1}{BM T_{\text{valid}}} \sum_{b,m,t} \mathbb{1}(t \in \text{valid}) \left\|\mathbf{p}_{t}^{\text{sim}} - \mathbf{p}_{t}^{\text{pred}}\right\|_2^2.
\end{equation}
where $\mathbf{p}_{t}^{\text{sim}}$ is the position derived from integrating the kinematic model, and $\mathbf{p}_{t}^{\text{pred}}$ is the model's direct prediction. This ensures trajectories respect physical motion constraints.

\subsubsection{Overall Loss and Training Strategy}

The total loss combines weighted components of all trajectory, collision, and dynamics losses:

\begin{equation}
\mathcal{L}_{\text{Total}} = \sum_{i} \lambda_i L_i,
\end{equation}

where \(\lambda_i\) are loss weights. We employ a curriculum learning strategy that initially optimizes base trajectory losses to ensure stable GT imitation, then progressively activates collision and dynamics losses in a post-warmup phase to enforce safety and physical realism without compromising training stability.

\section{Experiments}
\subsection{Experimental Setup}
We evaluate our framework on the Waymo Open Motion Dataset (WOMD) for prediction and planning tasks. All baseline models are retrained using default configurations with a fixed seed, as reproducing exact original results was not feasible.

\subsubsection{Implementation Details}
The encoder comprises 6 transformer layers with road maps represented as polylines (up to 20 points, $\sim$10m in WOMD). We select $N_m = 768$ nearest map polylines and $N_l = 192$ nearest reachable lanes around each agent. Local self-attention operates on 16 nearest neighbors with hidden dimension $D = 256$. The decoder stacks 6 layers with $L = 128$ dynamically selected map polylines for motion refinement. We employ 64 motion query pairs with command-specific intention points from k-means clustering partitioned by command type (left/right turn, straight, stationary) and globally for unknown and VRU categories. Non-maximum suppression selects the top 6 predictions from 64 trajectories using 2.5m endpoint distance threshold.

\subsubsection{Training Details}
We train PTR end-to-end using AdamW with learning rate = 0.0001, batch size 20, and weight decay = 0.01. Training spans 35 epochs on 4 H100 GPUs with learning rate decay (factor 0.5) every 2 epochs from epoch 20 to 30, plus 5 finetuning epochs. Loss weights: $\lambda_{\text{GMM, cls}} = 1.0, \lambda_{\text{dense, col, dynamic}} = 0.5$, determined via hyperparameter tuning. Collision and dynamics losses activate post-warmup (epoch 10) for training stability. We employ a teacher-student strategy where commands for predicted agents start with 90\% ground-truth availability and are progressively masked to 10\% "unknown" over the final 5 epochs, simulating inference conditions where surrounding agent intentions are unavailable.

\subsubsection{Prediction Task}
\label{subsec:pred_task}
The WOMD dataset supports marginal and joint prediction subtasks. Marginal prediction targets independent future trajectories for single agents, while joint prediction considers pairs of interacting agents. Each scenario provides 1 second of history and 8 seconds of predicted trajectories. Standard metrics namely minADE, minFDE, miss rate, overlap rate, and mAP~\cite{WAYMO2021} are used to assess trajectory accuracy and safety.

\subsubsection{Planning Task}
Planning evaluation is conducted on interactive WOMD scenarios~\cite{GameFormer2023, DIPP2023} using standard metrics~\cite{DIPP2023}: ADE and FDE measure prediction accuracy over the full horizon and at 5s; planning error, miss rate, and collision score assess planned trajectories exclusively. Planning error measures displacement deviations at horizons 1s, 3s, and 5s; miss rate measures spatial alignment with ground-truth trajectories via threshold-bounded regions; collision score quantifies safety by measuring collisions between planned and predicted trajectories. Predicted agents are the 10 closest to ego.

\subsection{Main Results}
\subsubsection{Marginal and Joint Prediction}

\begin{figure*}[ht]
    \centering
    \begin{tabular}{@{}c@{\hspace{-1.4cm}}l@{\hspace{2cm}}}
        \begin{minipage}[t]{\linewidth}
            % MTR Row
            \begin{subfigure}[b]{\linewidth}
                \centering
                \hspace{-1.5cm}
                \begin{minipage}[b]{0.04\linewidth}
                    \small
                    MTR\\
                    \\
                    \\
                    \\
                \end{minipage}
                \begin{subfigure}[b]{0.175\linewidth}
                    \includegraphics[width=\linewidth]{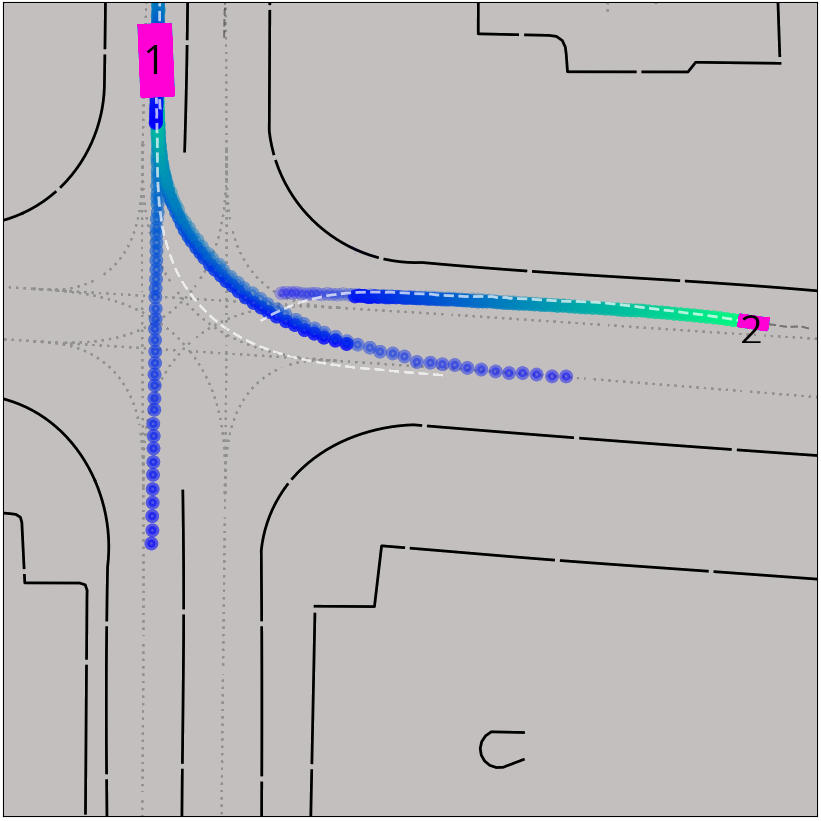}
                \end{subfigure}
                \hspace{-0.233cm}
                \begin{subfigure}[b]{0.175\linewidth}
                    \includegraphics[width=\linewidth]{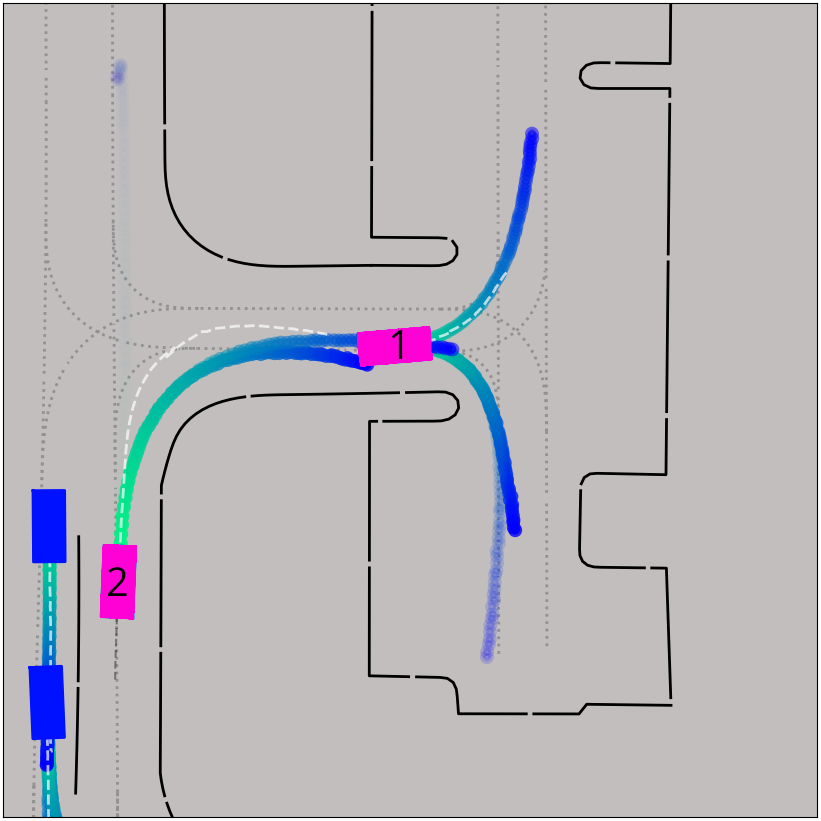}
                \end{subfigure}
                \hspace{-0.23cm}
                \begin{subfigure}[b]{0.175\linewidth}
                    \includegraphics[width=\linewidth]{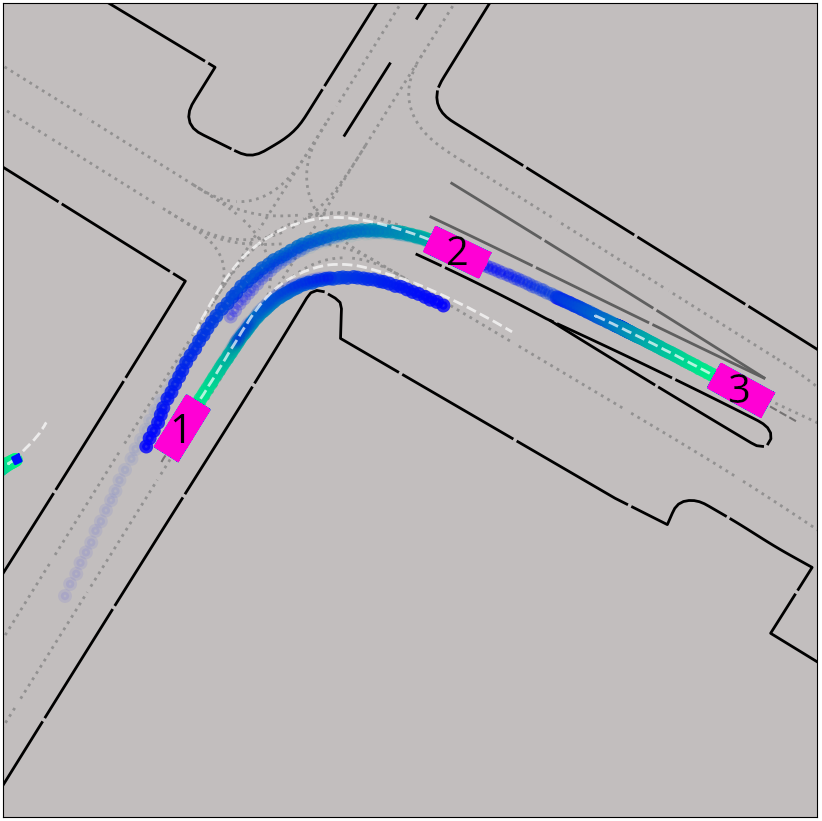}
                \end{subfigure}
                \hspace{-0.23cm}
                \begin{subfigure}[b]{0.175\linewidth}
                    \includegraphics[width=\linewidth]{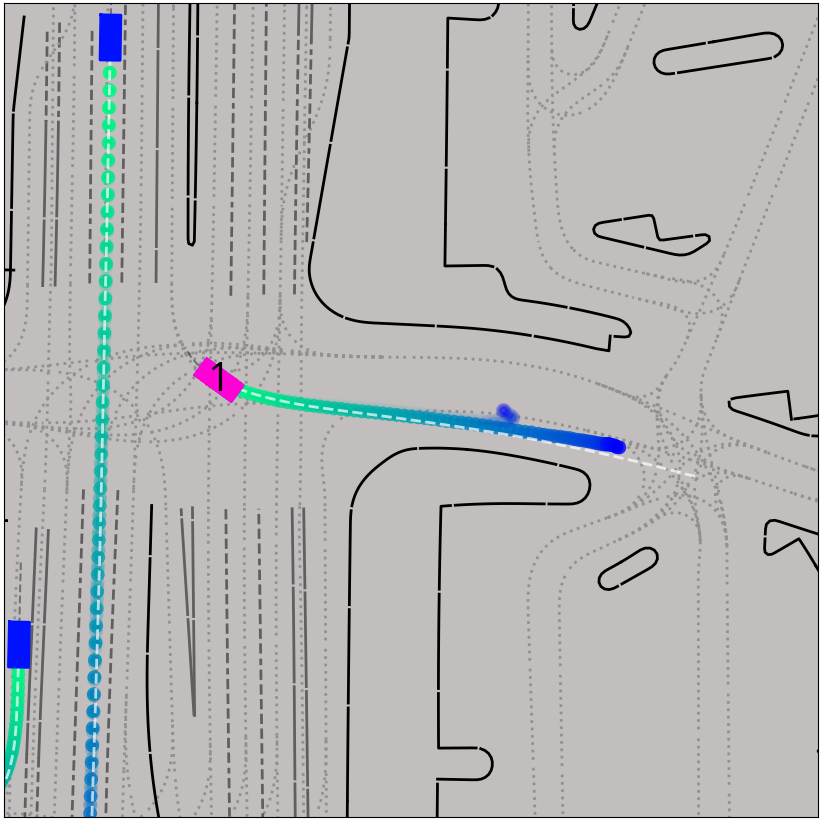}
                \end{subfigure}
                \hspace{-0.23cm}
                \begin{subfigure}[b]{0.175\linewidth}
                    \includegraphics[width=\linewidth]{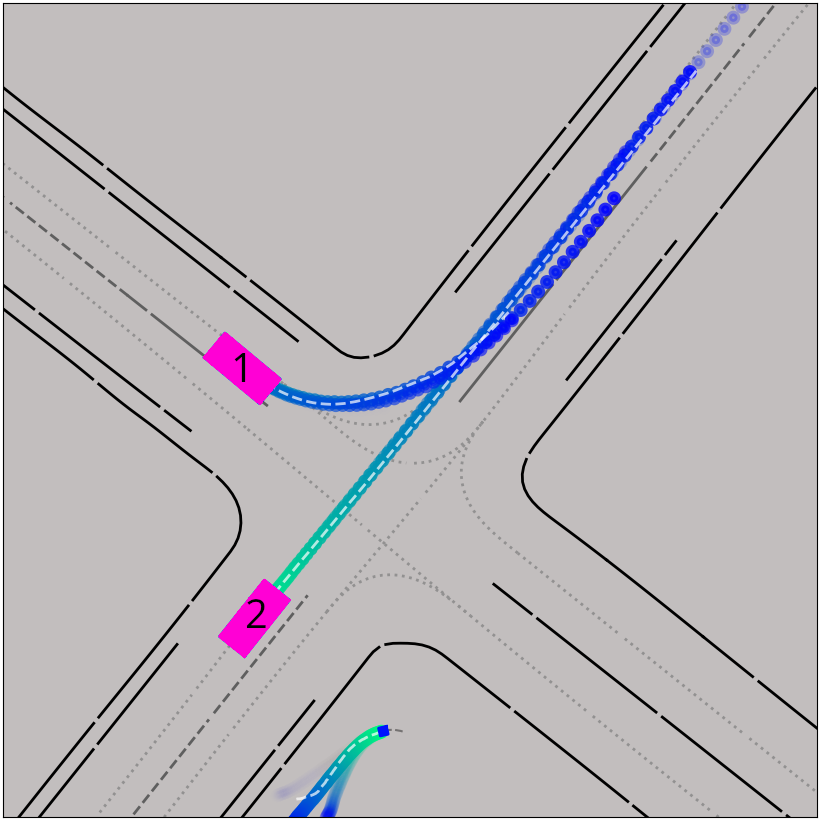}
                \end{subfigure}
            \end{subfigure}
            \\
            % PTR Row
            \begin{subfigure}[b]{\linewidth}
                \centering
                \hspace{-1.5cm}
                \begin{minipage}[b]{0.04\linewidth}
                    \small
                    Ours\\
                    \\
                    \\
                    \\
                    \\
                    \\
                \end{minipage}
                \begin{subfigure}[b]{0.175\linewidth}
                    \includegraphics[width=\linewidth]{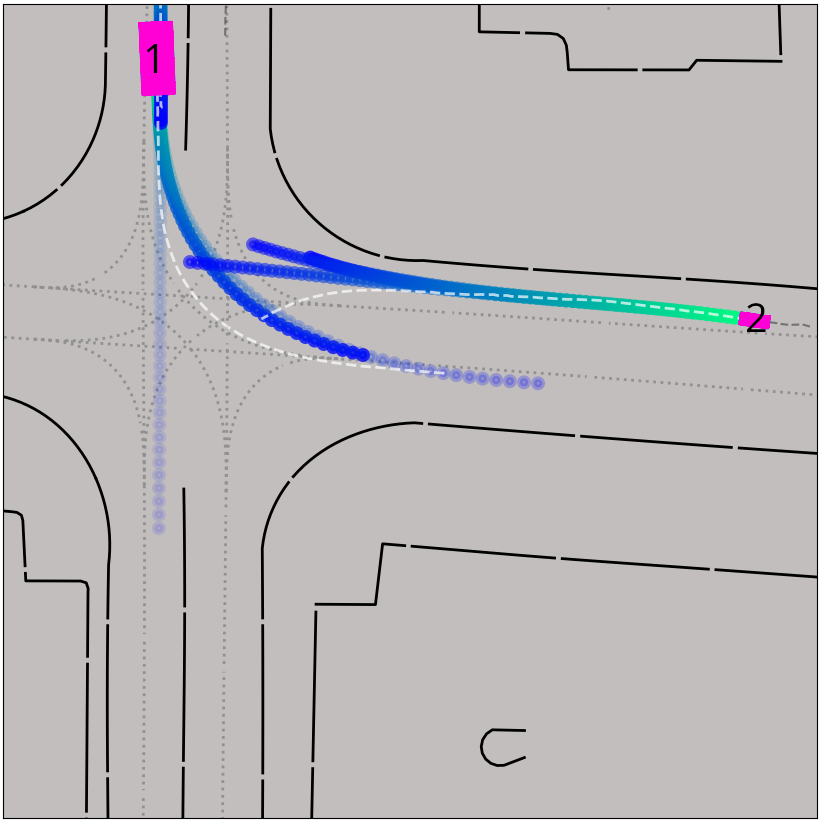}
                    \caption{Cautious cyclist}
                \end{subfigure}
                \hspace{-0.24cm}
                \begin{subfigure}[b]{0.175\linewidth}
                    \includegraphics[width=\linewidth]{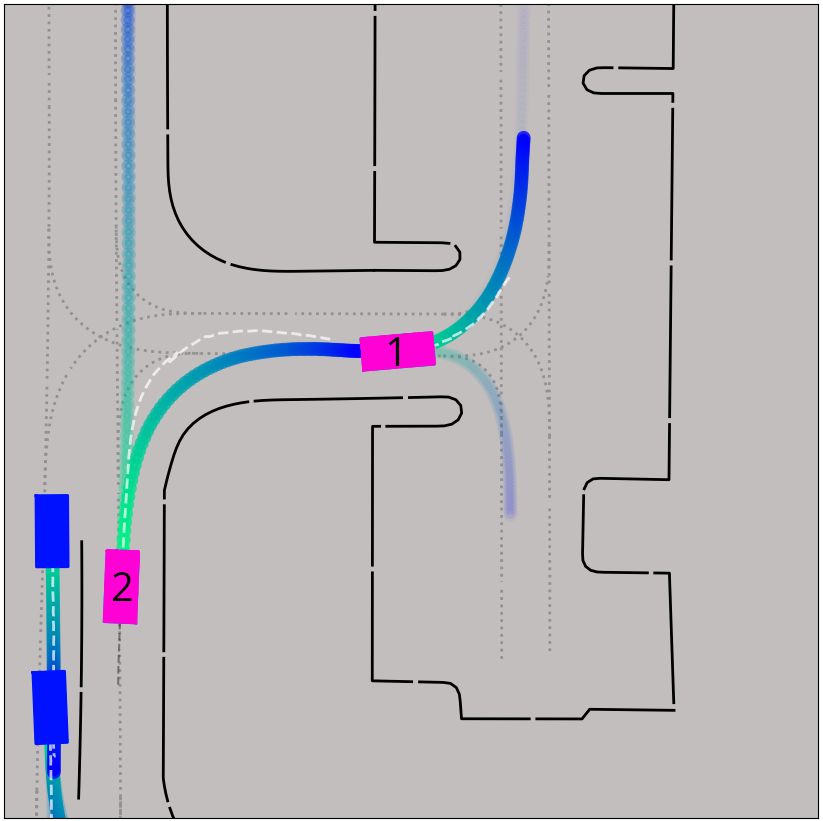}
                    \caption{Command guidance}
                \end{subfigure}
                \hspace{-0.23cm}
                \begin{subfigure}[b]{0.175\linewidth}
                    \includegraphics[width=\linewidth]{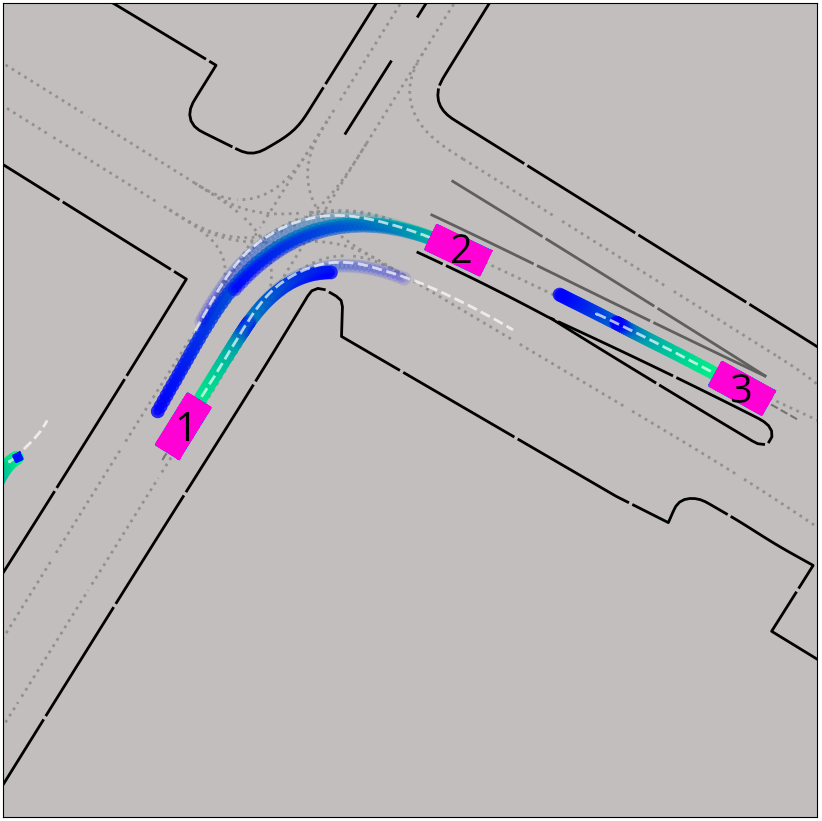}
                    \caption{Dense traffic}
                \end{subfigure}
                \hspace{-0.23cm}
                \begin{subfigure}[b]{0.175\linewidth}
                    \includegraphics[width=\linewidth]{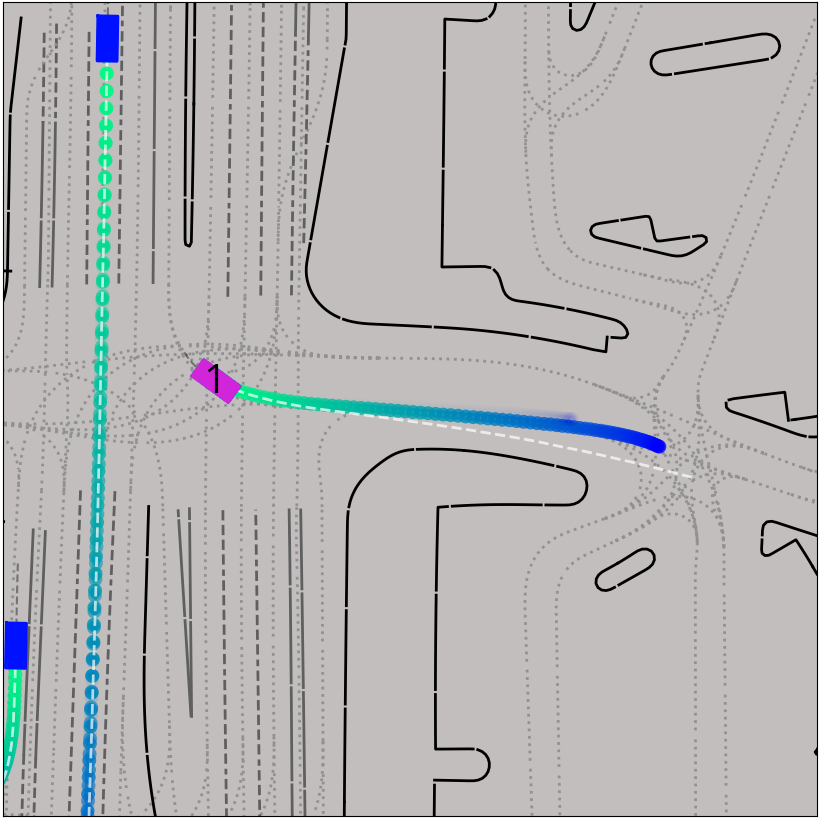}
                    \caption{Lane allignment}
                \end{subfigure}
                \hspace{-0.23cm}
                \begin{subfigure}[b]{0.175\linewidth}
                    \includegraphics[width=\linewidth]{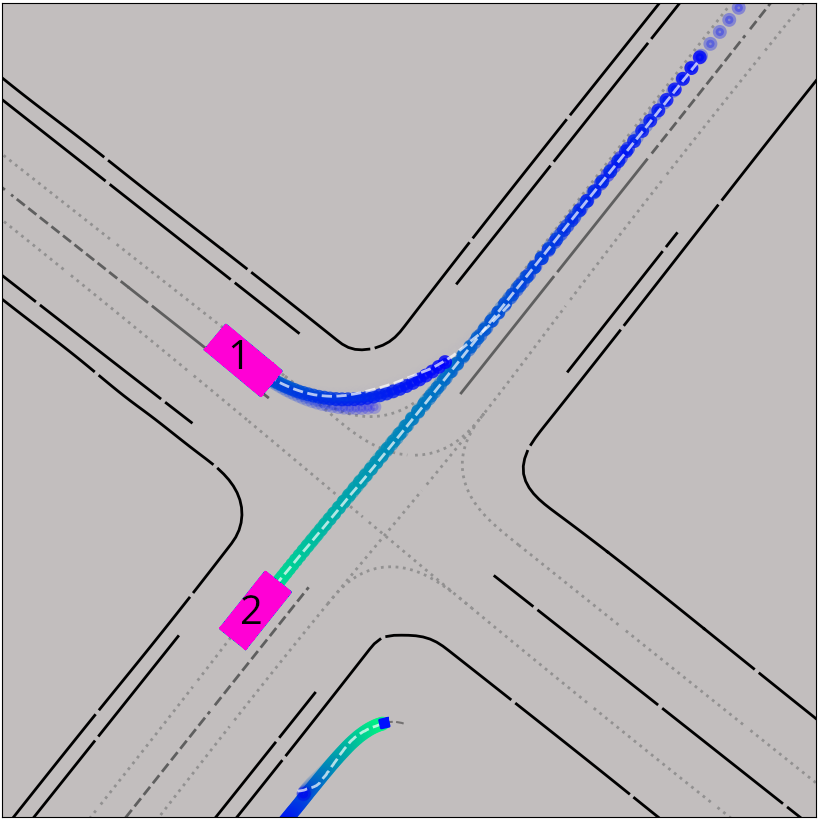}
                    \caption{Cautious merge}
                \end{subfigure}
            \end{subfigure}
        \end{minipage}
        &
        \hspace{0.1cm}
        \begin{minipage}[c]{0.04\linewidth}
            \includegraphics[width=1.8\linewidth, height=0.23\textheight]{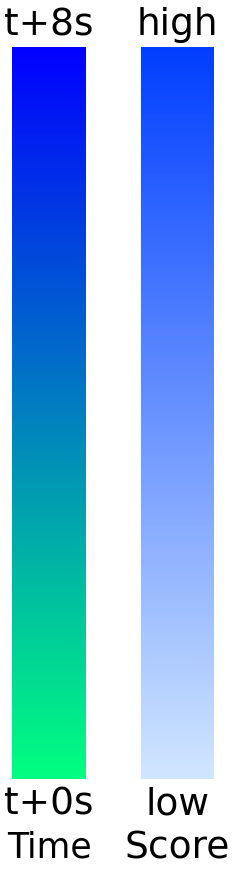}
        \end{minipage}
    \end{tabular}
    \caption{Qualitative top-3 prediction comparison of MTR (top) and PTR (bottom) across five scenarios. Scenario visualization shows agents of interest (pink boxes), other agents (blue boxes), prediction (blue lines), ground-truth (white dashed lines), lane centerline (gray dotted lines), lane separator (gray dashed lines), and road boarder (black solid lines). \textbf{a)} Vehicle (1) exhibits more confident left-turn predictions centered on lane markings in PTR; cyclist (2) has less GT alignment due to the near-miss interaction. \textbf{b)} PTR increases left-turn confidence for vehicle (1) through high-level command guidance, while vehicle (2) maintains multimodal predictions exploring alternative routes. \textbf{c)} PTR shows improved centerline adherence and GT alignment for vehicle (2); vehicles (1) and (3) generate more conservative trajectories due to collision constraints, deviating from GT endpoints for vehicle (1) while vehicle (3) deviates less. \textbf{d)} Vehicle (1) achieves better centerline alignment in PTR, improving map consistency but reducing GT fidelity in non-compliant driving behavior. \textbf{e)} PTR produces conservative trajectories for vehicle (1), while vehicle (2) continues nominal behavior without significant collision-avoidance influences.}
    \label{fig:prediction}
\end{figure*}

We evaluate both marginal and joint prediction performance on the top-6 predictions without assuming command knowledge, setting all high-level commands to "Unknown". For marginal prediction (Table~\ref{tab:marginal}) PTR demonstrates consistent improvements over MTR across all agent types: 4.3\% improvement in mAP, 1.5\% reduction in minADE (0.6023 vs. 0.6115), and 1.0\% reduction in minFDE (1.2325 vs. 1.2445), validating the effectiveness of goal-conditioned prediction in marginal evaluated scenarios.
For joint prediction (Table~\ref{tab:joint}), top-6 joint predictions are selected from 36 possible agent-pair combinations using confidence scores computed as the product of marginal probabilities. PTR achieves 3.5\% improvement in mAP and 1.0\% reduction in minADE (0.9470 vs. 0.9561) over MTR. Interestingly, minFDE increases slightly to 2.1956 from 2.1615 in the joint setting, which we attribute to the complexity of multi-agent interaction modeling and discuss further in Section~\ref{sec:discussion}. Figure~\ref{fig:prediction} provides qualitative examples demonstrating prediction quality across single-agent and interactive scenarios.

\begin{table}[ht]
\centering
\caption{Performance on the marginal validation set of WOMD
accessed across vehicle (V), pedestrian (P), cyclist (C) and
their average (AVG).}
\label{tab:marginal}
\begin{tabular}{l|c||cccc}
\hline
\rowcolor{gray!35}
method & type & mAP\textuparrow & minADE\textdownarrow & minFDE \textdownarrow & MR \textdownarrow \\
\hline
      & V    & 0.4357 & 0.7666 & 1.5444 & 0.1570 \\
MTR~\cite{MTR2022}   & P & 0.4092 & 0.3549 & 0.7453 & 0.0793 \\
      & C    & 0.3590 & 0.7131 & 1.4438 & 0.1853 \\
      & AVG        & 0.4013 & 0.6115 & 1.2445 & 0.1405 \\
\hline
      & V    & 0.4376 & 0.7601 & 1.5698 & 0.1553 \\
Ours   & P & 0.4243 & 0.3525 & 0.7335 & 0.0734 \\
      & C    & 0.3936 & 0.6943 & 1.3942 & 0.1809 \\
      & AVG        & \textbf{0.4185} & \textbf{0.6023} & \textbf{1.2325} & \textbf{0.1365} \\
\hline
\end{tabular}
\end{table}

\begin{table}[ht]
    \centering
    \caption{Performance on the joint validation set of WOMD accessed across vehicle (V), pedestrian (P), cyclist (C) and their average (AVG).}
    \label{tab:joint}
    \begin{tabular}{l|c||cccc}
        \hline
        \rowcolor{gray!35}
        method & type & mAP\textuparrow & minADE\textdownarrow & minFDE\textdownarrow & MR\textdownarrow \\
        \hline
              & V   & 0.2951  & 1.0126  & 2.2925  & 0.3911 \\
        MTR~\cite{MTR2022}   & P   & 0.2185  & 0.7505  & 1.6578  & 0.4098 \\
              & C   & 0.1156  & 1.1053  & 2.5342  & 0.5500 \\
              & AVG & 0.2097  & 0.9561  & \textbf{2.1615}  & 0.4503 \\
        \hline
              & V   & 0.3081 & 0.9562 & 2.2990 & 0.3881 \\
        Ours   & P   & 0.2222 & 0.7460 & 1.6388 & 0.4024 \\
              & C   & 0.1209  & 1.1388  & 2.6491  & 0.5554 \\
              & AVG & \textbf{0.2170} & \textbf{0.9470} & 2.1956 & \textbf{0.4486}\\
        \hline
    \end{tabular}
\end{table}

\subsubsection{Open-Loop Planning Evaluation}
\begin{figure*}[ht]
    \centering
    \begin{tabular}{@{}c@{\hspace{-0.5cm}}c@{\hspace{2cm}}}
        \begin{minipage}[t]{0.9\linewidth}
            % GameFormer Row
            \begin{subfigure}[b]{\linewidth}
                \centering
                \hspace{-0.5cm}
                \begin{minipage}[b]{0.07\linewidth}
                    \small
                    Game\\
                    Former\\
                    \\
                    \\
                \end{minipage}
                \begin{subfigure}[b]{0.19\linewidth}
                    \includegraphics[width=\linewidth]{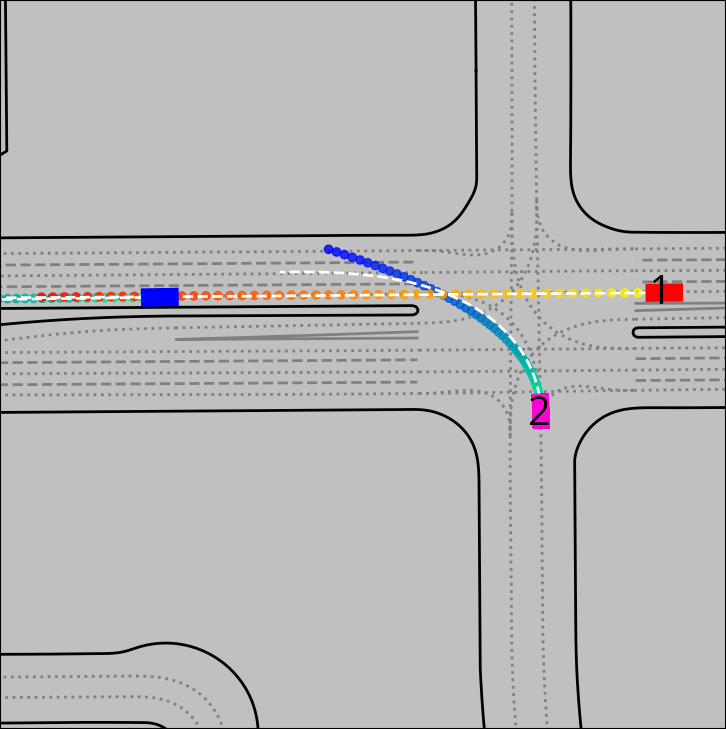}
                \end{subfigure}
                \hspace{-0.23cm}
                \begin{subfigure}[b]{0.19\linewidth}
                    \includegraphics[width=\linewidth]{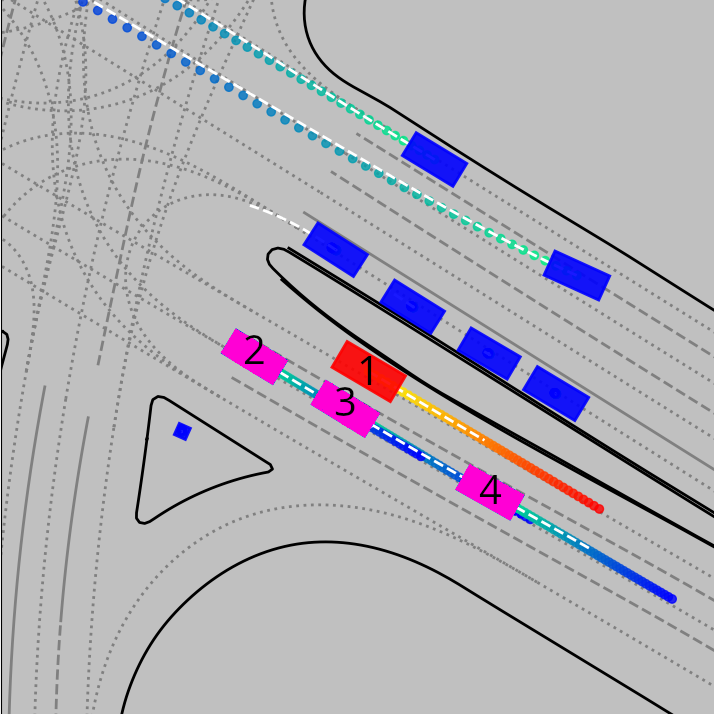}
                \end{subfigure}
                \hspace{-0.23cm}
                \begin{subfigure}[b]{0.19\linewidth}
                    \includegraphics[width=\linewidth]{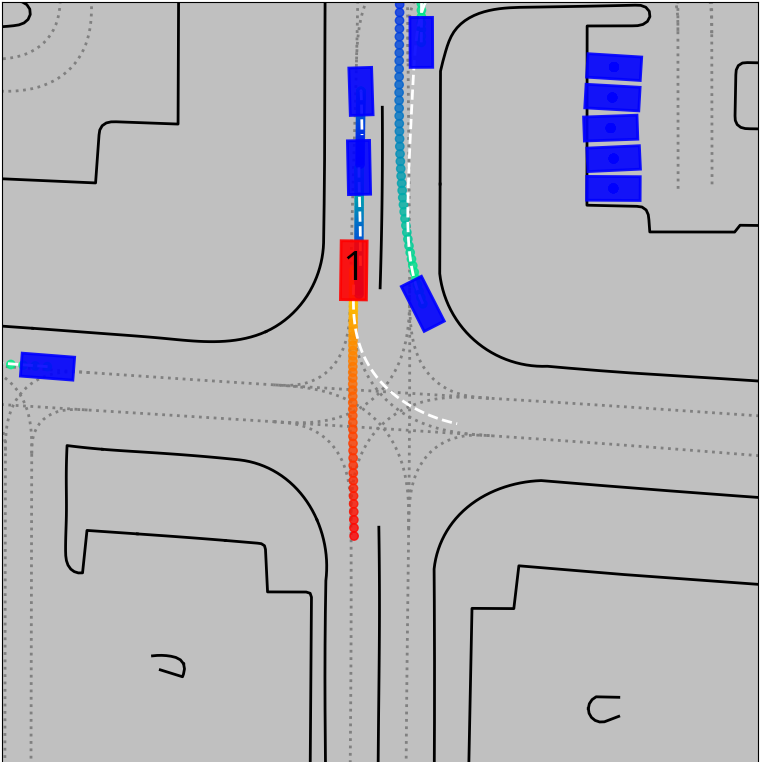}
                \end{subfigure}
                \hspace{0.7cm}
                \begin{subfigure}[b]{0.19\linewidth}
                    \includegraphics[width=\linewidth]{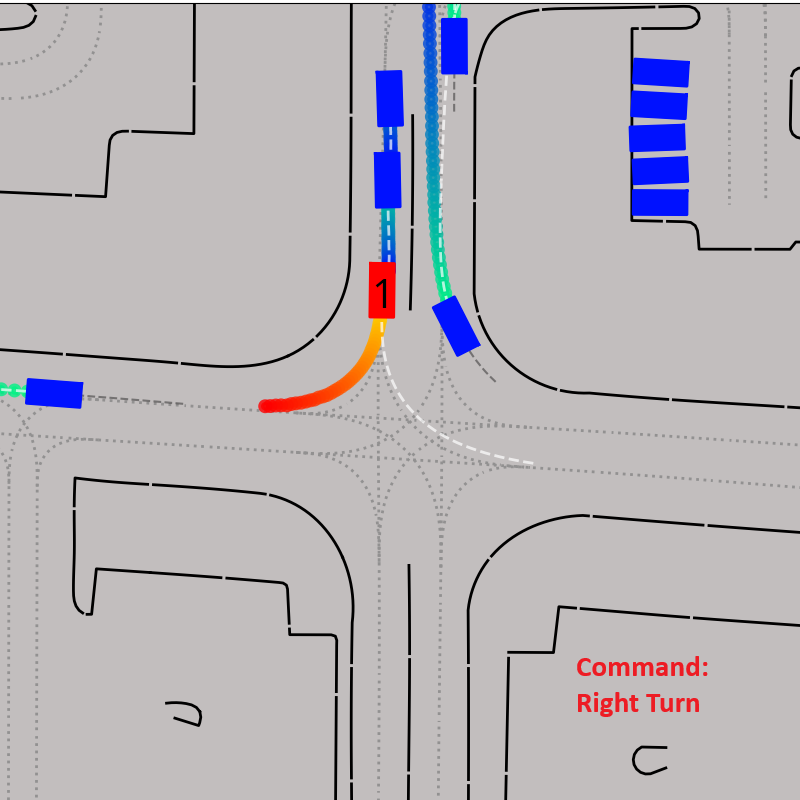}
                \end{subfigure}
            \end{subfigure}
            \\
            % PTR Row
            \begin{subfigure}[b]{\linewidth}
                \centering
                \hspace{-0.5cm}
                \begin{minipage}[b]{0.07\linewidth}
                    \small
                    Ours\\
                    \\
                    \\
                    \\
                \end{minipage}
                \begin{subfigure}[b]{0.19\linewidth}
                    \includegraphics[width=\linewidth]{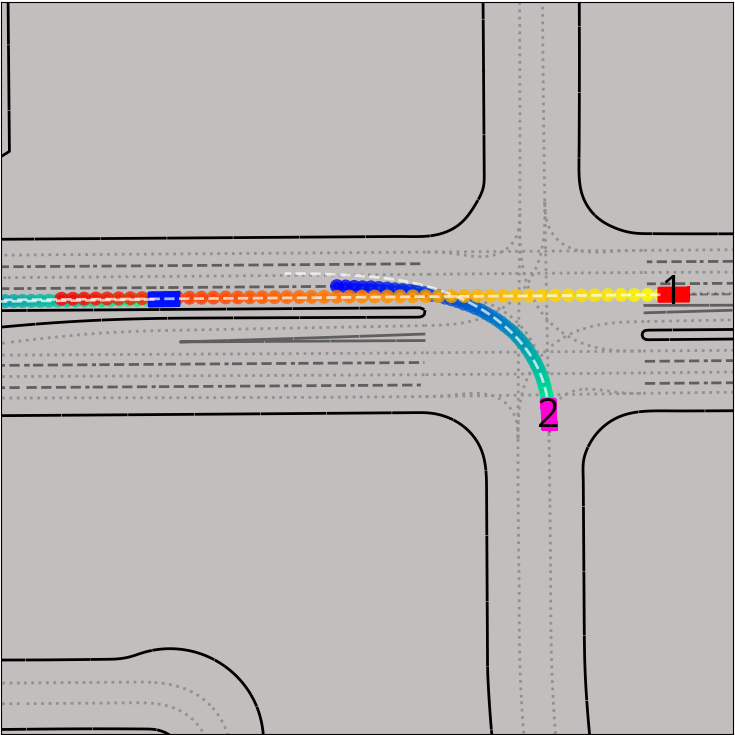}
                    \caption{Feasible path}
                \end{subfigure}
                \hspace{-0.23cm}
                \begin{subfigure}[b]{0.19\linewidth}
                    \includegraphics[width=\linewidth]{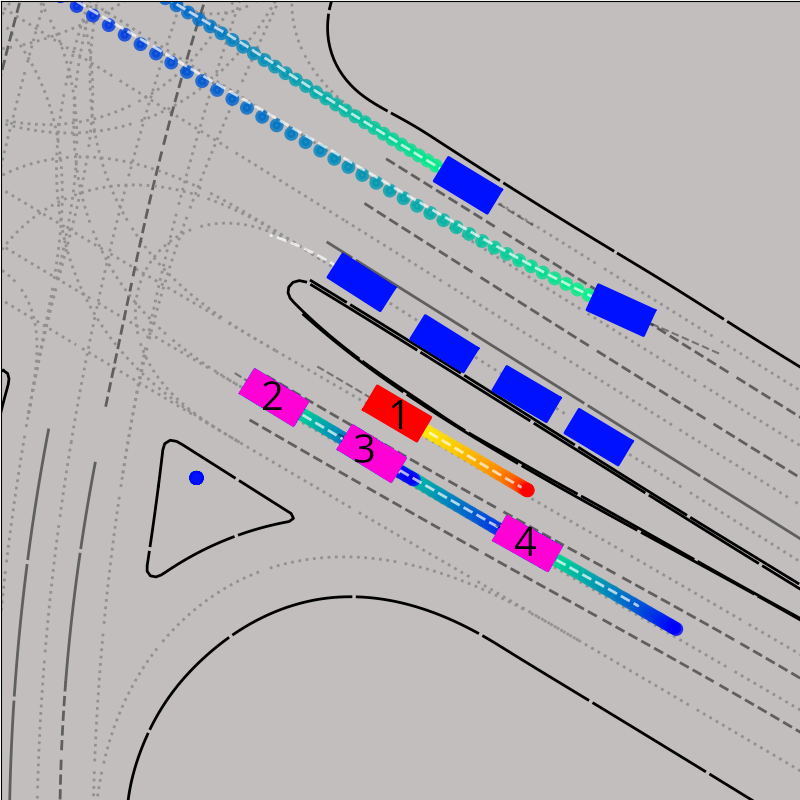}
                    \caption{Dense traffic}
                \end{subfigure}
                \hspace{-0.23cm}
                \begin{subfigure}[b]{0.19\linewidth}
                    \includegraphics[width=\linewidth]{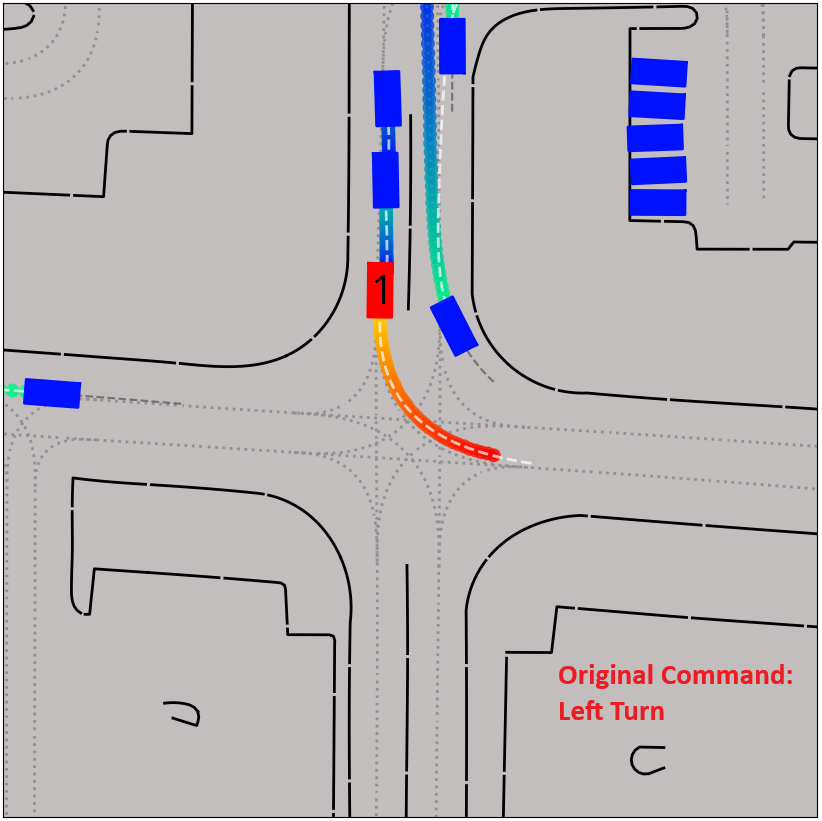}
                    \caption{Intention Allignment}
                \end{subfigure}
                \hspace{0.7cm}
                \begin{subfigure}[b]{0.19\linewidth}
                    \includegraphics[width=\linewidth]{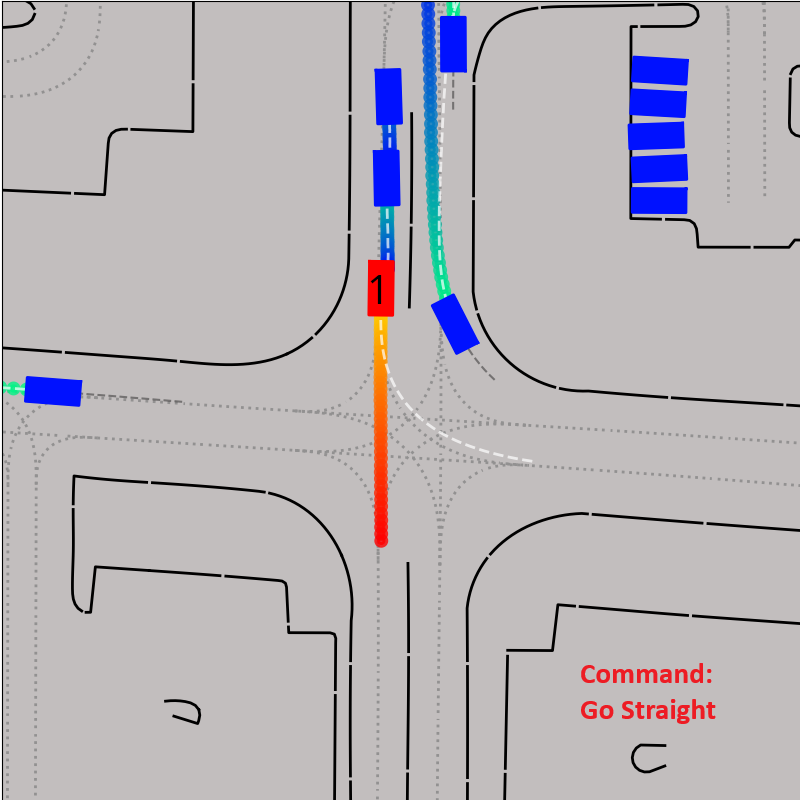}
                    \caption{Modified commands}
                \end{subfigure}
            \end{subfigure}
        \end{minipage}
        &
        \hspace{0.2cm}
        \begin{minipage}[c]{0.06\linewidth}
            \includegraphics[width=1.5\linewidth, height=0.25\textheight]{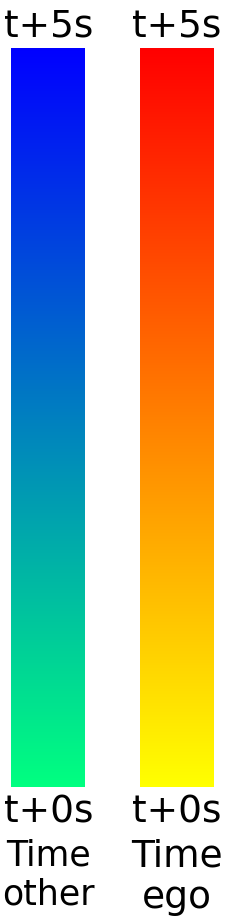}
        \end{minipage}
    \end{tabular}
    \caption{Qualitative top-1 planning comparison of GameFormer (top) and PTR (bottom) across three scenarios, demonstrating planning capabilities and high-level command influence (last pair). Scenario visualization shows ego (red box), agents of interest (pink boxes), other agents (blue boxes), prediction (blue lines), plan (red line), ground-truth (white dashed lines), lane centerline (gray dotted lines), lane separator (gray dashed lines), and road boarder (black solid lines). \textbf{a)} Ego (1) generates a similar trajectory as GameFormer, while surrounding vehicle (2) maintains a more feasible path. \textbf{b)} In a dense interactive scenario with nearby vehicles (2, 3, 4), ego (1) produces more conservative GT aligned planning trajectory in PTR compared to GameFormer. \textbf{c)} PTR's ego (1) planning aligns with the left-turn high-level command, while GameFormer's trajectory appears misaligned with the scenario intent. \textbf{d)} Same scenario as c) both images generated by PTR with modified commands (top: right turn; bottom: straight), demonstrating PTR's command-conditioned planning responsiveness to navigation guidance.}
    \label{fig:planning}
\end{figure*}

We evaluate PTR on the open-loop planning benchmark introduced by DIPP~\cite{DIPP2023} based on WOMD, following established protocols. Surrounding vehicle high-level commands are set to "Unknown" while the ego vehicle receives command guidance. Table~\ref{tab:plan_womd} presents the results. PTR achieves substantial improvements across safety and accuracy metrics: collision rate of 2.16\%, miss rate of 8.50\%, and planning error improvements of 5.9\% (0.117m), 4.5\% (0.835s), and 15.5\% (2.340s) at 1s, 3s, and 5s horizons, respectively, compared to the retrained GameFormer. Most notably, ADE and FDE improvements of 28.7\% (0.669) and 22.0\% (1.624) demonstrate substantial gains in prediction accuracy. These results validate that goal-conditioned guidance significantly improves both safety and planning fidelity in interactive scenarios. Figure~\ref{fig:planning} provides qualitative examples illustrating the effectiveness of command-driven trajectory generation in complex dynamic environments.

\begin{table*}
    \scriptsize
    \vspace{0.1cm}
    \centering
    \caption{Open-loop planning evaluation on WOMD, showing performance on ego and 10 closest predicted agents}
    \label{tab:plan_womd}
    \begin{tabular}{l||ccccc|cc}
        \hline
        \rowcolor{gray!35}
        Method & Collision Rate \textdownarrow & Miss Rate \textdownarrow & PE @1s \textdownarrow & PE @3s \textdownarrow & PE @5s \textdownarrow & ADE \textdownarrow & FDE \textdownarrow \\
        \hline
        Vanilla IL & 4.25 & 15.61 & 0.216 & 1.273 & 3.175 & -- & -- \\
        DIM~\cite{DIM2019} & 4.96 & 17.68 & 0.483 & 1.869 & 3.683 & -- & -- \\
        MultiPath++~\cite{MULTI2021} & 2.86 & 8.61 & 0.146 & 0.948 & 2.719 & -- & -- \\
        MTR-e2e~\cite{MTR2022} & 2.32 & 8.88 & 0.141 & 0.888 & 2.698 & -- & -- \\
        DIPP~\cite{DIPP2023} & 2.33 & 8.44 & 0.135 & 0.902 & 2.803 & 0.925 & 2.059 \\
        GameFormer~\cite{GameFormer2023} & \textbf{1.98} & \textbf{7.53} & 0.129 & 0.836 & 2.451 & 0.853 & 1.919 \\
        \hline
        GameFormer (retrained) & 2.71 & 12.18 & 0.124 & 0.873 & 2.702 & 0.861 & 1.982 \\
        \rowcolor{gray!20}
        Ours & 2.16 & 8.50 & \textbf{0.117} & \textbf{0.835} & \textbf{2.340} & \textbf{0.669} & \textbf{1.624} \\
        \hline
    \end{tabular}
\end{table*}

\subsection{Ablation Study}
\label{sec:ablation}
We evaluate component contributions using 20\% uniformly sampled frames (approximately 97k scenes) from WOMD training data, preserving original distribution. All models use marginal prediction metrics on WOMD validation set.

\subsubsection{Component Integration Analysis}
As shown in Table~\ref{tab:ablation_1}, high-level commands alone provide notable improvements in mAP, demonstrating that semantic guidance directly benefits mode-ranking quality. However, without complementary constraints, positional accuracy remains suboptimal, indicating that command guidance must be paired with consistency mechanisms. Adding Dynamics Loss substantially improves trajectory plausibility, reducing minADE and Miss Rate while maintaining mAP gains. Collision Loss further reduces overlap rate while sacrificing displacement metrics slightly, reflecting the conservative nature of collision avoidance. Incorporating Reachable Lanes recovers displacement performance while preserving safety improvements, indicating that route constraints effectively reconcile safety and accuracy objectives. The full model achieves the improved mAP in Table~\ref{tab:ablation_1}, demonstrating the synergistic effect of all components, though with minor trade-offs compared to intermediate configurations.

\subsubsection{Goal-Conditioning Integration Strategy}
We compare three command integration strategies: (1) One-Hot Concatenation appends command vectors to agent features; (2) MLP Fusion processes command embeddings through dimension-expanding MLPs; (3) Decoder Query Preset (proposed) initializes decoder queries with command embeddings.

The one-hot approach achieves strong displacement metrics but lower mAP, indicating early fusion guides trajectory endpoints without effectively constraining prediction confidence. MLP fusion further degrades performance through feature dimension bottlenecking (\(2D \to D\) projection), which negates command semantics. The proposed preset approach directly initializes query content features, providing explicit guidance during trajectory generation. This decoupling improves mAP by +1.26\% relative to one-hot while incurring minADE and minFDE increases (+0.006m and +0.134m respectively), reflecting a favorable trade-off between endpoint accuracy and mode-ranking quality.

\begin{table*}
    \scriptsize
    \vspace{0.1cm}
    \centering
    \caption{Ablation study on proposed modules, showing how each component contributes to performance improvement}
    \begin{tabular}{>{\centering\arraybackslash}p{1.2cm}| >{\centering\arraybackslash}p{1.4cm} >{\centering\arraybackslash}p{1.4cm} >{\centering\arraybackslash}p{1.4cm} >{\centering\arraybackslash}p{1.4cm}||>{\centering\arraybackslash}p{1.15cm} >{\centering\arraybackslash}p{1.1cm}  >{\centering\arraybackslash}p{1.1cm} >{\centering\arraybackslash}p{1.25cm} >{\centering\arraybackslash}p{1.1cm}}
        \hline
        \rowcolor{gray!40}
        Method & High-Level Command & Dynamics Loss & Collision-Loss &  Reachable Lanes  & minADE \textdownarrow & minFDE \textdownarrow & Miss Rate \textdownarrow & Overlap Rate \textdownarrow  & mAP \textuparrow \\
        \hline
        MTR & $\times$ & $\times$ & $\times$ & $\times$ & 0.6695 & 1.3776 & 0.1653 & 0.0434 & 0.3469\\
        \hline
         & \checkmark & $\times$ & $\times$ & $\times$ & 0.6727 & 1.3744 & 0.1676 & 0.0444 & 0.3529 \\
         & \checkmark & \checkmark & $\times$ & $\times$ &  \textbf{0.6680s} & \textbf{1.3732} & \textbf{0.1645} & 0.0441 & 0.3536 \\
        Ours & \checkmark & \checkmark & \checkmark & $\times$ &  0.6787 & 1.4040 & 0.1677 & \textbf{0.0431} & \textbf{0.3550} \\
         & \checkmark & \checkmark & \checkmark & \checkmark & 0.6681 & 1.3770 & 0.1656 & 0.0433  & 0.3543 \\
         \hline
    \end{tabular}
    \label{tab:ablation_1}
\end{table*}

\begin{table}[h]
    \scriptsize
    \vspace{0.1cm}
    \caption{Ablation Study on High-Level Command Incorporation into Transformer Decoder}
    \centering
    \begin{tabular}{>{\centering\arraybackslash}p{0.9cm}||>{\centering\arraybackslash}p{1.15cm} >{\centering\arraybackslash}p{1.1cm}  >{\centering\arraybackslash}p{1.3cm} >{\centering\arraybackslash}p{1.1cm}}
        \hline
        \rowcolor{gray!40}
        Variation & minADE \textdownarrow & minFDE \textdownarrow & Miss Rate \textdownarrow  & mAP \textuparrow \\
        \hline
        One-Hot & \textbf{0.6666} & \textbf{1.3610} & \textbf{0.1649} & 0.3485 \\
        MLP &  0.6706 & 1.3635 & 0.1661 & 0.3447  \\
        Preset & 0.6727 & 1.3744 & 0.1676 & \textbf{0.3529}\\
        \hline
        
    \end{tabular}
    \label{tab:ablation_2}
\end{table}

\section{Discussion}
\label{sec:discussion}
The results demonstrate complementary strengths and trade-offs among the proposed components. High-level commands enable semantically-guided prediction without explicit command supervision at inference. Reachable lanes constrain predictions to valid routes, improving feasibility. Dynamics loss ensures kinematic consistency. However, notable limitations exist.
Collision loss improves safety-awareness but prioritizes collision avoidance, leading to conservative predictions and increased minFDE in joint scenarios despite mAP improvements. Collision constraints encourage deviations from ground-truth endpoints to avoid interactions (Figure~\ref{fig:prediction}, a), and the planning module may amplify this conservatism by generating unnecessarily evasive maneuvers. In dense multi-agent scenarios (Figure~\ref{fig:planning}, b), PTR generates more conservative, ground-truth aligned trajectories compared to GameFormer, highlighting the safety-fidelity trade-off: collision avoidance improves safety metrics at the cost of trajectory realism.
Route compliance constraints can fail when agents exhibit rule-noncompliant behavior such as illegal lane changes or u-turns, missing plausible future trajectories in real-world scenarios. Command-conditioning improves semantic alignment (Figure~\ref{fig:planning}, c) and demonstrates strong responsiveness to guidance (Figure~\ref{fig:planning}, d); however, this benefit diminishes under command uncertainty or agent non-compliance. Balancing safety constraints with behavioral realism remains an open challenge. This difficulty stems from an 'imitation-safety gap' where aggressive human maneuvers conflict with our absolute priority on safety. Increased displacement metrics reflect a deliberate choice to favor collision avoidance over pure imitation. Future research into context-aware weighting could enable the model to adaptively modulate caution by reducing speeds and increasing safety buffers based on traffic complexity. This approach ensures that safety remains the primary objective in every scenario.

\section{Conclusion}
We present PTR, a unified framework addressing the prediction-planning gap in autonomous driving. By integrating goal-conditioned prediction, dynamic feasibility, collision avoidance, and lane-level topology within a Transformer, PTR jointly optimizes prediction and planning. A teacher-student strategy progressively masks agent commands, aligning training with inference conditions. On the Waymo Open Motion Dataset, PTR achieves 4.3\%/3.5\% improvements in marginal/joint mAP over MTR and 15.5\% planning error reduction at 5 seconds versus GameFormer. The architecture-agnostic design supports application to diverse Transformer-based models. Future work includes enriching the command taxonomy with finer-grained actions (merging, yielding, lane maintenance) and learning context-conditioned behavioral refinements for complex scenarios. Additionally, integration and further validation on the nuPlan benchmark~\cite{NuPlan2021} to further validate its generalization across diverse urban environments and closed-loop scenarios and on a real vehicle are planned, which is being developed within the Project STADT:up.

\section*{ACKNOWLEDGMENT}
This work is a result of the joint research project STADT:up (19A22006N). The project is supported by the German Federal Ministry for Economic Affairs and Climate Action (BMWK), based on a decision of the German Bundestag. The author is solely responsible for the content of this publication.

\end{document}